\documentclass[sigconf]{acmart}
\usepackage{booktabs} %
\usepackage{verbatim}
\usepackage{CJKutf8}
\usepackage{makecell}
\usepackage[utf8]{inputenc}
\usepackage{algorithm}
\usepackage{multirow}
\usepackage{caption}
\usepackage{subcaption}
\usepackage{amsmath,amsfonts,amsthm,bm}
\usepackage{algorithmic}
\usepackage{algorithm}  
\usepackage{algorithmic,eqparbox,array}

\AtBeginDocument{%
  \providecommand\BibTeX{{%
    \normalfont B\kern-0.5em{\scshape i\kern-0.25em b}\kern-0.8em\TeX}}}

\setcopyright{acmcopyright}
\copyrightyear{2022} 
\acmYear{2022} 
\setcopyright{acmcopyright}\acmConference[KDD '22]{Proceedings of the 28th ACM SIGKDD Conference on Knowledge Discovery and Data Mining}{August 14--18, 2022}{Washington, DC, USA}
\acmBooktitle{Proceedings of the 28th ACM SIGKDD Conference on Knowledge Discovery and Data Mining (KDD '22), August 14--18, 2022, Washington, DC, USA}
\acmPrice{15.00}
\acmDOI{10.1145/3534678.3539086}
\acmISBN{978-1-4503-9385-0/22/08}

\begin{document}

\title{No One Left Behind: Inclusive Federated Learning over Heterogeneous Devices}

\author{
Ruixuan Liu
}
\affiliation{
   \institution{
    School of Information, Renmin University of China
   }
   \city{Beijing}
   \postcode{100872}
   \country{China}
}
\email{ruixuan.liu@ruc.edu.cn}

\author{Fangzhao Wu}
\authornote{The corresponding author.}
\affiliation{%
  \institution{Microsoft Research Asia}
    \city{Beijing}
    \postcode{100080}
  \country{China}
}
\email{wufangzhao@gmail.com}

\author{Chuhan Wu}
\affiliation{%
  \institution{Department of Electronic Engineering, Tsinghua University}
    \city{Beijing}
    \postcode{100084}
  \country{China}
}
\email{wuchuhan15@gmail.com}

\author{Yanlin Wang}
\affiliation{%
  \institution{Microsoft Research Asia}
    \city{Beijing}
    \postcode{100080}
  \country{China}
}
\email{yanlwang@microsoft.com}

\author{Lingjuan Lyu}
\affiliation{%
  \institution{Sony AI}
    \city{Tokyo}
    \postcode{xxx}
  \country{Japan}
}
\email{lingjuan.lv@sony.com}

\author{Hong Chen}
\affiliation{
   \institution{
    School of Information, Renmin University of China
   }
   \city{Beijing}
   \postcode{100872}
   \country{China}
}
\email{chong@ruc.edu.cn}

\author{Xing Xie}
\affiliation{%
  \institution{Microsoft Research Asia}
    \city{Beijing}
    \postcode{100080}
  \country{China}
}
\email{xingx@microsoft.com}

\renewcommand{\shortauthors}{Ruixuan Liu et al.}

\begin{abstract}
Federated learning (FL) is an important paradigm for training global models from decentralized data in a privacy-preserving way. 
Existing FL methods usually assume the global model can be trained on any participating client. 
However, in real applications, the devices of clients are usually heterogeneous, and have different computing power. 
Although big models like BERT have achieved huge success in AI, it is difficult to apply them to heterogeneous FL with weak clients.
The straightforward solutions like removing the weak clients or using a small model to fit all clients will lead to some problems, such as under-representation of dropped clients and inferior accuracy due to data loss or limited model representation ability.
In this work, we propose InclusiveFL, a client-inclusive federated learning method to handle this problem. 
The core idea of InclusiveFL is to assign models of different sizes to clients with different computing capabilities, bigger models for powerful clients and smaller ones for weak clients.
We also propose an effective method to share the knowledge among local models with different sizes.
In this way, all the clients can participate in FL training, and the final model can be big and powerful enough. 
Besides, we propose a momentum knowledge distillation method to better transfer knowledge in big models on powerful clients to the small models on weak clients. 
Extensive experiments on many real-world benchmark datasets demonstrate the effectiveness of InclusiveFL in learning accurate models from clients with heterogeneous devices under the FL framework.
\end{abstract}



\maketitle
\section{Introduction}
Nowadays, huge amounts of data are generated every day from a wide range of devices such as mobile phones, autonomous vehicles, or servers~\cite{surveyFlmcmahan2021advances}.
Exchanging data from different sources to train machine learning models is attractive but challenging under the privacy regularization such as GDPR\footnote{\url{https://gdpr-info.eu/}}.
Based on the principles of focused collection and data minimization to avoid privacy issues, federated learning (FL)~\cite{mcmahan2017communication} emerges as a privacy-aware paradigm to train a global model while keeping decentralized data stay on local devices.
In each round, sampled clients pull a copy of the global model and optimize models in parallel on local datasets.
Then, the server aggregates local models to update the global model.
Hence, conventional federated learning methods~\cite{mcmahan2017communication, reddi2020adaptive} make an essential assumption that all clients have sufficient local resources to train models with the same architecture.

\begin{figure}
    \centering
    \includegraphics[trim=200 200 250 150, clip, width=0.45\textwidth]{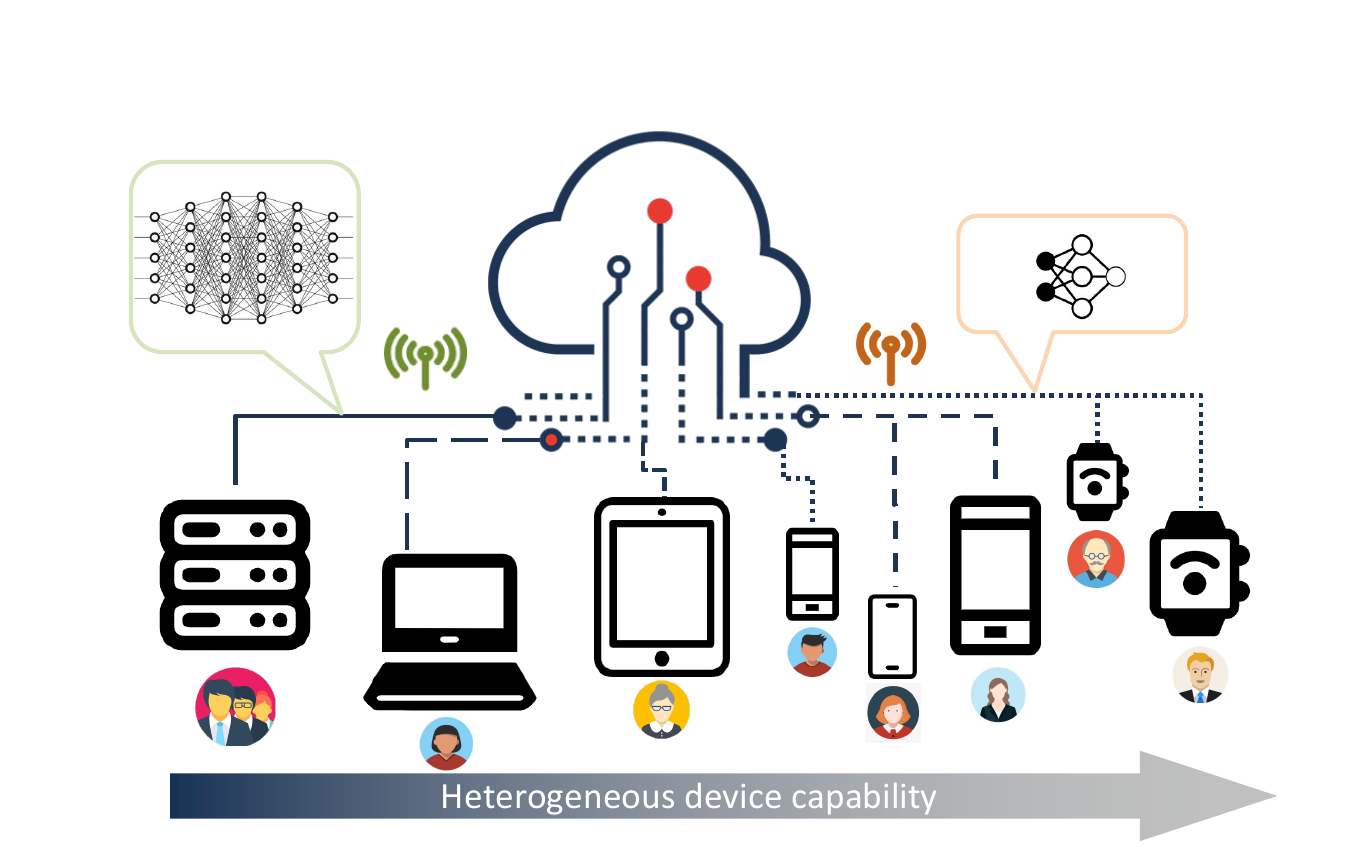}
    \caption{InclusiveFL over heterogeneous devices.}
    \label{fig-teaser}
\end{figure}

However, as shown in Fig. \ref{fig-teaser}, in the real world, devices of clients are usually heterogeneous and they may have the significantly different computing power and memory size~\cite{varghese2016challenges,xu2021asynchronous, hou2020dynabert}.
Although a large global model (e.g., BERT with 12-layer Transformer~\cite{bert, roberta2019}) is promising to achieve a good performance, it is difficult to apply it in heterogeneous FL because weak clients that are constrained by hardware resources cannot afford the cost of local training.
A straightforward way~\cite{bonawitz2019towards} is to drop weak clients and only aggregate parameters from powerful clients with sufficient resources.
But excluding data on weak devices in training leads to fairness issues~\cite{HaoELZLCC21, mehrabi2021survey} because weak clients would be underrepresented in the final global model.
Furthermore, the loss of data on weak clients causes an inferior accuracy~\cite{joseph2015}, especially when the amount of weak clients is enormous (e.g., in less developed areas).
Alternatively, a client-inclusive baseline is choosing a small global model to fit the minimum capability that all clients can offer.
However, the representation ability of the final global model would be largely limited by the small model architecture~\cite{liu2018rethinking}.

Due to the limitations of the above methods, it is attractive to include all clients with heterogeneous devices while utilizing the capabilities of large models.
Thus, an intuitive way is training bigger local models on powerful clients and smaller local models on weak clients.
To exchange knowledge over heterogeneous models, several works~\cite{li2019fedmd,chang2019cronus, sun2020federated} utilize the knowledge distillation technique~\cite{hinton2015distilling} to enhance the global model with an ensemble of local predictions.
However, they assume clients share a public proxy dataset, which is impractical for weak clients with a limited memory.
A recent work \textit{HeteroFL}~\cite{diao2020heterofl} does not require a proxy dataset and proposes to train heterogeneous local models by averaging the overlapped sub-matrices of all parameter matrices across various models.
However, this method is not beneficial for retaining large models' knowledge in smaller models because the width pruning operation breaks the original model structure.
In addition, even the same sub-matrix may have different behaviors in the small and large models due to the model structure differences, which may lead to a suboptimal performance due to the mismatch of feature spaces.
Moreover, simply sharing parts of parameters across different models cannot effectively transfer useful knowledge encoded by strong models to other weaker models~\cite{hinton2015distilling}.
Thus, how to learn models with different sizes and aligned feature spaces via effective knowledge transfer is a challenging problem in FL with heterogeneous devices.

In this paper, we propose a client-inclusive federated training solution \textit{InclusiveFL} that can train a large model over devices with heterogeneous capabilities, and address the above key challenges.
\textit{InclusiveFL} assigns models of different sizes to clients with different computing capabilities, bigger models for powerful clients and smaller ones for weak clients.
To eliminate the mismatch between small and large models, we propose to share shallow bottom model layers in the largest model with other smaller models.
It has been empirically investigated that the lower layers are most critical for retaining the pre-trained knowledge in downstream tasks \cite{sajjad2020effect}, and the feature spaces among different layers have much consistency~\cite{dodge2020fine}.
We propose a layer-wise heterogeneous aggregation method to update the parameters of shared bottom layers. 
Furthermore, since the small model on weak clients may not be strong enough, we propose to transfer the knowledge learned by large models on powerful clients to small models on weak clients via a momentum distillation method~\cite{wu2021newsbert}.
Intuitively, by encouraging the top encoder layer in small models to imitate the behavior of top encoder layers in a larger model, we can achieve effective knowledge transfer in \textit{InclusiveFL}.
Contributions of this paper are summarized as follows:
\begin{itemize}
    \item We propose \textit{InclusiveFL} for effective federated learning over heterogeneous client devices.
    Clients with different computing capabilities are assigned models of different sizes, and all of them can contribute to learning a large and powerful global model.
    \item We propose an effective method to share the knowledge in the heterogeneous models from heterogeneous clients. 
    Besides, we propose a momentum knowledge distillation method to boost knowledge sharing from large models to small models.
    \item We conduct extensive experiments to verify the effectiveness of the proposed method in learning an accurate global model from heterogeneous clients in the FL framework. 
\end{itemize}
\section{Background and Related Work}
Federated learning~\cite{mcmahan2017communication} is a machine learning technique that trains an algorithm across multiple decentralized edge devices (cross-device)~\cite{hard2018federated, mcmahan2018learning, yang2018applied} or servers holding local data samples (cross-silo)~\cite{warnat2021swarm, courtiol2019deep, wu2022fedctr}, without exchanging them.
Conventional federated learning protocol typically adopts \textit{FedAvg}~\cite{mcmahan2017communication} to aggregate local parameters as follows:
$$
w_{t+1} \leftarrow \sum_{k=1}^K \frac{n_k}{n} w_{t+1}^k,
$$
where $n=\sum_{k=1}^K n_k$ denotes the total number of local samples.
For a better training convergence, \textit{FedAdam}~\cite{reddi2020adaptive} is recently proposed to use adaptive methods on the server and SGD on the clients.
For each round, a sampled client $i$ computes the local update $g_t^i$ after iterating over several rounds of local SGDs.
Then, the server averages local updates $g_t \leftarrow \frac{1}{|\mathcal{S}|}\sum_{i\in |\mathcal{S}|} g_t^i$ and updates the global model as follows:
\begin{equation*}
    \begin{split}
        m_t \leftarrow & \beta_1 m_{t-1} + (1-\beta_1) g_t \\
        v_t \leftarrow & \beta_2 v_{t-1} + (1-\beta_2) g_t^2 \\
        w_{t+1} \leftarrow & w_t + \eta \frac{m_t}{\sqrt{v_t} + \tau}. \\
    \end{split}
\end{equation*}

Basic federated learning methods make an essential assumption that all clients have sufficient local resources to train models with the same architecture.
However, weak clients cannot even compute over the global model when local hardware is constrained.
Previous works~\cite{li2019fedmd,chang2019cronus} provide solutions for federated learning with heterogeneous local models based on knowledge distillation\cite{hinton2015distilling}.
A server student model is updated by distilling from the ensemble of predictions~\cite{lin2020ensemble}.
Alternatively, local models are optimized sequentially~\cite{li2019fedmd} or jointly~\cite{chang2019cronus} over the augmented dataset and the local private dataset.
However, it is hard to guarantee the domain and quality of the public data. 
And it is impractical for weak clients with limited hardware to store the public dataset.

Recently, \textit{HeteroFL}~\cite{diao2020heterofl} removes this assumption and conduct parameter-averaging over heterogeneous models.
Authors scale model architectures for clients with variant local resources with variant widths.
For each matrix $W\in\mathbb{R}^{d_g\times k_g}$ in the large model, they reduce the parameters for a smaller model as the upper left sub-matrix $W_s=W[:d_s, :k_s]$ with size $d_s\times k_s$.
To exchange knowledge, heterogeneous models share the overlapped matrix with a parameter averaging.
However, \textit{HeteroFL} has limitations of a biased initial pruning and parameter mismatch across heterogeneous models.
Instead, we reduce number of layers for small models and propose methods to solve these problems.

Other related works that consider weak capabilities include solutions to reduce the communicational cost~\cite{li2019fedmd,sattler2019robust}.
Several works aim to mitigate the up-link communication bottlenecks from client to server\cite{lin2020deep, wangni2018gradient, dutta2020discrepancy, konevcny2016federated, wang2018atomo}.
Local model updates can be sparsified, quantized or randomly sub-sampled before uploading to the central server.
\citet{konevcny2016federated} performs a lossy compression on the model updates and reduces the error incurred by the subsequent quantization with the randomized Hadamard transform.
To reduce the server-to-client communication, \citet{caldas2018expanding} extends the idea of dropout~\cite{dropout} and drops a fixed number of activation units at each full-connected layer before distributing a local model to each client.
\citet{bouacida2021adaptive} adopts an adaptive strategy and selects the dropped activation based on the activation score map.
It should be noted that local optimization is conducted on decompressed models.
Also, above works do not consider federated training over heterogeneous models and are compatible to our solution when aggregating local models with the same architecture in \textit{InclusiveFL}.
\section{Methodology}
In this section, we introduce our solution \textit{InclusiveFL}, which trains a global %
model across client devices with heterogeneous capabilities.
We will first introduce the layer-wise heterogeneous aggregation framework in \textit{InclusiveFL}, and then introduce details of our proposed momentum distillation in federated learning for more effective knowledge exchange in the heterogeneous setting.

\begin{figure*}
    \centering
    \includegraphics[trim=100 90 100 130,clip, width=\textwidth]{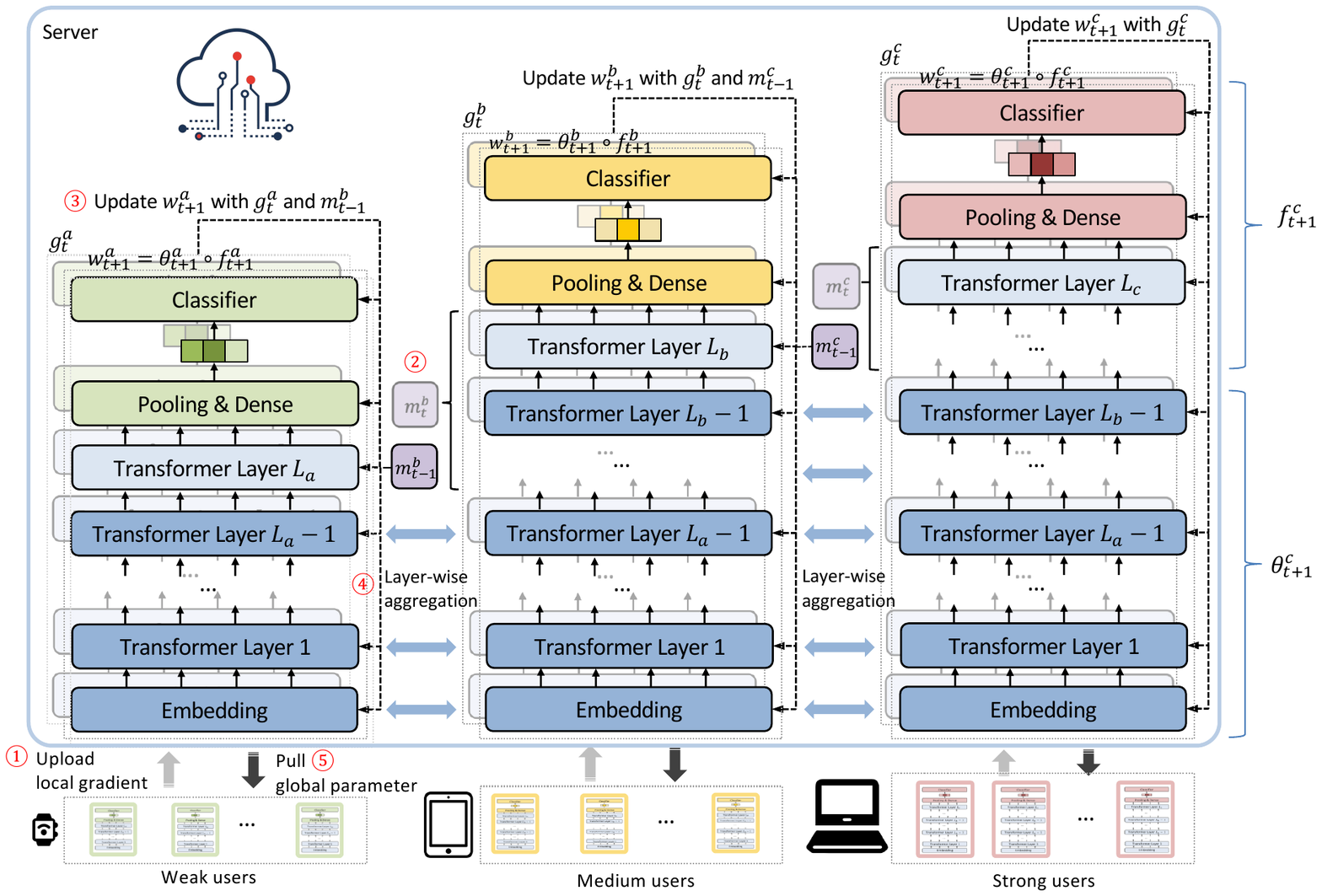}
    \caption{Overview of \textit{InclusiveFL}.}
    \label{fig-overview}
\end{figure*}

\subsection{Layer-wise Heterogeneous Aggregation Framework}
The overview of \textit{InclusiveFL} framework is shown in Fig. \ref{fig-overview}.
Without loss of generality, we denote three groups that hold different device capabilities as weak, medium, and strong clients.
Our core idea is to assign models of different sizes to clients with different computing capabilities, bigger models for strong clients and smaller models for weak clients.
We reduce the number of parameters for weaker clients by decreasing the depth of network.
Since the bottom layers are most critical for retraining the pre-trained knowledge~\cite{sajjad2020effect},  the sub-model with pre-trained bottom layers would benefit from the pre-trained parameters, which results in better initialization for small models.
Accordingly, the clients in weak, medium or strong client groups own small, medium or a large local models with $L_a$, $L_b$, $L_c$ layers, where $L_a<L_b<L_c$.
In general, we conduct a homomorphic aggregation within the same client group and denote the aggregated model as the sub-global models $w^a$, $w^b$, and $w^c$.
Then we run a heterogeneous aggregation across different client groups to update all sub-global models.

\begin{algorithm}[t]
	\small
	\caption{Framework of \textit{InclusiveFL}.} \label{alg-training} \label{alg-InclusiveFL}
	\begin{algorithmic}[1]
	    \REQUIRE
	   \STATE Initialize $\theta_0^c, f_0^a, f_0^b, f_0^c, m_0^b\leftarrow0, m_0^c\leftarrow 0$
	   \STATE Set shared layers $\theta_0^a \leftarrow \theta_0^{c, 1:L_a}, \theta_0^b \leftarrow \theta_0^{c, 1:L_b}$
	   \STATE Get models $w_0^a \leftarrow \theta_0^a \circ f_0^a, w_0^b \leftarrow \theta_0^b \circ f_0^b, w_0^c \leftarrow \theta_0^c \circ f_0^c$
	   \FOR {$t \in \{1, \cdots, T\}$}
	      \STATE $U_t \leftarrow $ (randomly sample a portion of $r$ users)
	      \STATE $U_t^a, U_t^b, U_t^c \leftarrow$ (group $U_t$ into the same device type)
	      \STATE $n_a, n_b, n_c \leftarrow $ (the number of users for each group)
	      \STATE $//$\textbf{ Homomorphic aggregation}
	      \FOR {device type $j \in \{a, b, c\}$ }
	        \FOR {user $i\in U_t^{j}$}
	            \STATE Pull the sub-global model $w_t^j$
	            \STATE $g_t^{i, j} \leftarrow \text{LocalUpdate}(w_t^j, D_i)$ \quad $ \triangleright D_i$ is the local dataset
	        \ENDFOR
	        \STATE $g_t^j \leftarrow \frac{1}{|U_t^j|} \sum_{i\in U_t^j} g_t^{i, j}$
	        \IF {$j$ is not $c$} \label{line-mom-start}
	            \STATE $k \leftarrow $ (the device type that is greater than $j$ )
	            \STATE $g_t^{j, L_j} \leftarrow \beta \cdot m_{t-1}^k + (1-\beta) \cdot g_t^{j, L_j}$
	        \ENDIF
	        \STATE $w_{t+1}^j \leftarrow \text{FedAdam}(g_t^j, w_t^j)$ \quad $\triangleright w_{t+1}^j$ includes $ \theta_{t+1}^j \circ f_{t+1}^j $
	        \IF {$j$ is not $a$}
	            \STATE $L^- \leftarrow $ (the number of layers in the next smaller model)
	            \STATE $m_t^j \leftarrow \frac{1}{L_j - L^- + 1} \cdot \sum_{l \in [L^-:L_j]} g_t^{j, l}$ 
	        \ENDIF \label{line-mom-end}
	      \ENDFOR
	      \STATE $//$ \textbf{Heterogeneous aggregation} \label{line-hete-start}
	      \STATE $\theta_{t+1}^{c, l} \leftarrow \sum_{i\in \{a, b, c\}} \frac{n_i}{n_a + n_b + n_c} \theta_{t+1}^{i, l}$ for $l \in [1, \cdots, L_a-1]$ \label{line-share1}
	      \STATE $\theta_{t+1}^{c, l} \leftarrow \sum_{i \in \{b, c\}} \frac{n_i}{n_b+n_c} \theta_{t+1}^{i, l}$ for $l \in [L_a, \cdots, L_b-1]$ 
	      \label{line-share2}
	      \STATE Update $\theta_{t+1}^a \leftarrow \theta_{t+1}^{c, 1:L_a-1} \circ \theta_{t+1}^{a, L_a}, \theta_{t+1}^b \leftarrow \theta_{t+1}^{c, 1:L_b-1} \circ \theta_{t+1}^{b, L_b}$
	   \STATE $w_{t+1}^a \leftarrow \theta_{t+1}^a \circ f_{t+1}^a, w_{t+1}^b \leftarrow \theta_{t+1}^b \circ f_{t+1}^b, w_{t+1}^c \leftarrow \theta_{t+1}^c \circ f_{t+1}^c$ \label{line-comb}
	   \ENDFOR
	\end{algorithmic}
\end{algorithm}

As shown in Algorithm \ref{alg-InclusiveFL}, in the $t^{th}$ round of communication, the server first randomly samples a subset of clients with heterogeneous devices.
The sampled clients are grouped by the device type as $U_t^a, U_t^b$ and $U_t^c$ and the proportion $n_a:n_b:n_c$ for each device type is dynamic for each round.
Then, for a client $i$ in the group of device $j$, we conduct the local optimization with SGD as in conventional FL~\cite{mcmahan2017communication} and get the local update as $g_t^{i, j}$.
Within each client group, we average the update as $g_t^j$, where $j\in\{a, b, c\}$.
Then we update the sub-global model into $w_{t+1}^j$ with the averaged updates and the momentum $m_{t-1}^k$ to merge local knowledge from clients with the same device type.
The momentum $m_{t-1}^k$ is distilled from a larger model on client group $k$ and we explain details in Section \ref{sec-mom}.

With sub-models $w_{t+1}^a, w_{t+1}^b, w_{t+1}^c$ after the homomorphic aggregation over local models with an identical architecture, we start the heterogeneous aggregation from Line \ref{line-hete-start} in Algorithm \ref{alg-InclusiveFL} to exchange knowledge across variant model architectures.
It should be noted that we decompose the model $w$ into a shared module $\theta$ and an isolated module $f$ for each group.
The shared module includes bottom encoding layers $\theta$ and the isolated module includes the last encoder layer and top task-specific layers, such as the pooling and dense layer and the classifier layer in Fig. \ref{fig-overview}.
\textit{InclusiveFL} is a general framework for other tasks by only changing the architecture of the top layers $f$.
Thus, we first exclude the updated $f_{t+1}$ in the heterogeneous parameter aggregation and conduct the standard \textit{FedAvg} for each overlapped layer in shared module $\theta_{t+1}^c$, as shown in Line \ref{line-share1} and \ref{line-share2}.
Specifically, we use the notation $\theta_{t+1}^{i, l}$ to indicate the $l^{th}$ encoder layer in the model for device type $j \in \{a, b, c\}$.
Similarly, layers from $1$ to $L_a-1$ in the sub-global model with type $c$ is denoted as $\theta_{t+1}^{c, 1:L_a-1}$.
Combining with the within-group updated top layers $f_{t+1}$ for $\{a, b, c\}$ in Line \ref{line-comb}, the server derives the updated sub-global model and distributes to clients for the next round.

Intuitively, the bottom layers can be shared because they can capture basic features across heterogeneous models.
Also, the overlapped layers across three models have the same initialization, which help to align them keep aligned from the beginning of training.
We are motivated to isolate the last encoder layer and top layers for two considerations.
First, isolating top layers help mitigate the parameter mismatch in each sub-global model. 
For the example of a text classification task in Fig. \ref{fig-overview}, the bottom $L_a - 1$ layers in $w^a$ will produce a sequence of hidden states as the input to a pooling and dense layer for getting a sentence embedding.
However, the output of the bottom $L_a-1$ layers in $w_b$ can be further extracted with other $L_b-L_a+1$ layers before the pooling and dense layer.
Thus, the top parameters behave differently due to the heterogeneous model architectures.
If sharing $f_{t+1}^a$ with $f_{t+1}^b$ and $f_{t+1}^c$, the averaged model applied to $w^a$ would lack several layers of feature extraction, which would lead to a poor performance of $w_a$.
Second, isolating top layers makes \textit{InclusiveFL} capable of training separate local model with different target labels. 
For example in the cross-silo scenario such as federated named entity recognition~\cite{fedner}, medical data stored at different medical institutions may have different characteristics and annotation criterion, which leads to personalized top layers.

\subsection{Momentum Distillation}\label{sec-mom}
Now we introduce details of the momentum distillation for accelerating the heterogeneous federated training, as shown from Line \ref{line-mom-start} to \ref{line-mom-end} in Algorithm \ref{alg-InclusiveFL}.
Intuitively, a larger model is more capable of extracting knowledge from data than a small model.
Thus, we are motivated to distill knowledge from larger models to smaller models.
The knowledge distillation \cite{hinton2015distilling} in the centralized setting typically uses a hidden loss to align hidden representations of a small student model and a large teacher model.
However, the hidden loss does not apply to the heterogeneous federated learning with data separated for the small model and teacher model.

Instead, we notice that the key to align small models and large models when bottom layers are shared is to align the behaviors of the isolated encoder layer in a small model to the corresponding top layers in the larger model.
Hence, taking the small model and the medium model for instance, we calculate a gradient momentum as the average over updates for a stack of layers from $L_a$ to $L_b$ in the medium sub-global model.
For the averaged update $g_t^a$, we inject a distilled gradient momentum $m_{t-1}^b$ to the update $g_t^{a, L_a}$ of the isolated encoder layer $L_a$, which helps the $L_a$ layer in the small model imitate the block of layer $L_a$ to $L_b$ in the medium model.
Similarly, we use the averaged update from the large model as the gradient momentum for the layer $L_b$ in the medium model.
In general, the distillation transfers knowledge from a larger model to a smaller model without the limitation on the number of model types.
The gradient momentum is initialized as zero and applied with a momentum factor $\beta$, which is an important hyper-parameter to tune the strength of the distillation.

\section{Experiments}
In this section, we evaluate \textit{InclusiveFL} for federated learning over heterogeneous devices on IID and non-IID datasets and compare with both naive baselines and HeteroFL \cite{diao2020heterofl}.
To simulate the heterogeneous device scenario, we set up three types of clients: weak clients, medium clients and powerful clients.
Accordingly, they can store and compute over 4-layer, 8-layer and 12-layer transformer models, respectively.

\textbf{Datasets.}
Our experiments are conducted on the text classification task in the GLUE benchmark and token classification task (NER) on three medical corpora.
GLUE benchmark \cite{glue2018} is a collection of multiple natural language understanding tasks, including 
MNLI \cite{mnli_2018} (inference), 
SST-2 \cite{sst2_2013} (sentimental analysis), 
MRPC \cite{mrpc2005} (paraphrase detection), 
CoLA \cite{cola2019} (linguistic acceptability), 
QNLI \cite{qnli2018} (inference), QQP\footnote{\url{https://quoradata.quora.com/First-Quora-Dataset-Release-Question-Pairs}} (question-answering), 
RTE~\footnote{\url{https://huggingface.co/datasets/glue/viewer/rte/train}} (inference), 
and STSB \cite{stsb2017} (textual similarity).
The generality makes the GLUE dataset a standard benchmark to evaluate NLU models.

Three medical datasets includes SMM4H \cite{smm4h2019}, CADEC \cite{cadec2015} and ADE \cite{ade2012}, which are widely used to evaluate NER tasks.
The statistics of the medical datasets are shown in Table \ref{tab-dataset}.

\begin{table}[h]
\centering
\small
\caption{Statistics of the medical NER datasets.}\label{tab-dataset}
\begin{tabular}{@{}l|l|l|l@{}}
\toprule
\textbf{Dataset}                & \textbf{\# Sentences} & \textbf{Entity Types}                                                           & \textbf{\# Entity} \\ \midrule
SMM4H & 3,824 & ADE (1,707) & 1,707 \\ \midrule
ADE              & 4,483         & \begin{tabular}[c]{@{}l@{}}ADE (5,678), Drug (5,076), \\ Dosage (222)\end{tabular}           & 10,976     \\ \midrule
${\rm CADEC}$ & 7,683         & \begin{tabular}[c]{@{}l@{}}ADE (5,937), 
	Drug(1,796), \\ 
	Disease(282), Finding(425),\\ 
	Symptom(268)
	\end{tabular} & 8,535      \\ 
\bottomrule
\end{tabular}
\end{table}

\begin{table*}[t]
\centering\caption{Performance comparison of federated learning methods on the GLUE benchmark for text classification tasks. Performance is evaluated on the final large model in a global inference scenario.
Bold faces indicate the best method for inclusive FL training with heterogeneous devices.
}\label{tab-glue}
\begin{tabular}{c|c|cccccccc|c}
\toprule[1.0pt]
   & Inclusive?    & COLA  & MNLI  & MRPC  & QNLI  & QQP   & RTE   & SST2  & STSB  & Avg.  \\ \hline
All-Large   & N/A               & 63.03 & 86.48 & 91.50  & 92.09 & 91.49 & 76.12 & 94.43 & 90.60  & 85.72 \\ \hline
ExclusiveFL  & No               & 37.77 & 85.98 & 89.87 & 91.24 & 89.47 & 62.17 & 94.06 & 89.26 & 79.98 \\
All-Small   & Yes               & 34.91 & 78.83 & 82.50  & 85.93 & 79.37 & 58.94 & 90.14 & 83.68 & 74.29 \\ \hline
HeteroFL  & Yes               & 8.15  & 31.83 & 81.51 & 62.70  & 73.79 & 52.71 & 84.98 & 30.54 & 53.28 \\ \hline
InclusiveFL   & Yes                & \textbf{54.85} & \textbf{86.36} & \textbf{91.42} & \textbf{91.76} & \textbf{90.55} & \textbf{66.14} & \textbf{94.17} & \textbf{89.94} & \textbf{83.15} \\ 
\bottomrule[1.0pt]
\end{tabular}
\end{table*}

\begin{table*}[t]
\caption{Performance comparison of federated learning methods on medical datasets for named entity recognition.
Performances are evaluated for each named entity on each local model (SMM4H with 4-layer model, ADE with 9-layer model and CADEC with 12-layer model) in the local inference scenario.
Best method for heterogeneous training is shown in bold face.
}
\label{tab-ner}
\begin{tabular}{c|c|c|ccc|ccccc|c}
\toprule[1.0pt]
\multirow{2}{*}{Methods} & \multirow{2}{*}{Inclusive?} & 
SMM4H & 
\multicolumn{3}{c|}{ADE} & \multicolumn{5}{c|}{CADEC} &
\multirow{2}{*}{Avg.} \\ 
\cline{3-11} &  & ADE   & Drug   & ADE     & Dose   & ADE   & Symptom & Drug  & Disease & Finding & \\ 
\cline{1-12}
AllLarge                 & N/A & 55.08 & 95.21  & 80.91  & 18.97 & 71.23 & 43.37   & 90.43 & 33.55   & 29.85   & 57.62                \\ \hline
Local                    & No & 46.77 & 95.12  & 80.14  & 11.25 & 44.15 & 19.90    & 53.53 & 20.58   & 16.53   & 43.11                \\
AllSmall                 & Yes & 49.56 & 92.76  & 75.67  & 13.51 & 64.82 & 26.71   & 87.91 & 21.07   & 20.23   & 40.89                \\ \hline
HeteroFL                 & Yes & 27.94 & 73.90   & 37.57  & 11.57 & 59.90  & 26.92   & 84.58 & 25.77   & 19.88   & 42.51                \\ \hline
InclusiveFL                  & Yes & \textbf{49.90}  & \textbf{95.49}  & \textbf{80.34}  & \textbf{13.91}  & \textbf{71.10}  & \textbf{40.91}   & \textbf{90.21} & \textbf{39.34}   & \textbf{22.97}    & \textbf{56.02}                \\ \bottomrule[1.0pt]
\end{tabular}
\end{table*}

\textbf{Evaluation Protocols.}
For the evaluation on cross-device \cite{surveyFlmcmahan2021advances} federated learning with IID data, we randomly split the training dataset of each GLUE benchmark to 1000 clients and report the best performance of the evaluation dataset.
By default, we report the matched accuracy for MNLI, the Matthew's correlation for CoLA, Pearson correlation for STSB, F1-score for MRPC and the accuracy metric for all other tasks.
Higher value indicates better performance.
To evaluate over non-IID data distribution, we take each medical corpus as the local dataset of one client (e.g. a hospital) and simulate the cross-silo \cite{surveyFlmcmahan2021advances} federated setting.
We set SMM4H as the local data for a weak client, ADE as the local data for a medium client and CADEC as the local data for the powerful client.
We validate the generality of this setting by rotating datasets and client types and draw similar observations.
For each client, we randomly take 80\% of the local dataset as the training dataset and the rest as the evaluation dataset.
We use the seqeval framework\footnote{\url{https://github.com/chakki-works/seqeval}} to report the maximum averaged F1-score, precision, recall and accuracy on the evaluation dataset to indicate the overall performance or the performance over each type of entity for each client.
Results for the two settings are reported by averaging over 5 independent repeats.

\textbf{Baselines.}
We compare our proposed \textit{InclusiveFL} solution with the following naive baselines: 
\begin{itemize}
\item  \textit{AllLarge}: an ideal baseline to indicate the performance upper bound of training over heterogeneous devices. 
We remove the resource limitation on weak and medium clients and conduct federated training over all clients participant with a 12-layer RoBERTa model.
\item \textit{AllSmall}: a naive baseline to include all clients with heterogeneous devices and conduct training over the smallest model with 4 layers.
\item \textit{ExclusiveFL}: a naive baseline where weak and medium clients are dropped off in federated training due to the resource limitation on the largest model.
\item \textit{Local}: a naive baseline for cross-silo setting where each client trains the model with the size that fits their local resources.
\item \textit{HeteroFL}~\cite{diao2020heterofl}: the most related work for training across all clients over heterogeneous devices.
The number of parameters in smaller models are reduced in width by cropping each parameter matrix of the largest model into sub-matrix with various reduction radio.
\end{itemize}

\textbf{Hyper-Parameter Settings.}
For federated training over heterogeneous devices, we assign each client a client type with equal probability before training and fix it during training by default.
The effect of different ratios of device types is investigated in our hyper-parameter analysis.
For a fair comparison, we keep the number of parameters for each model type in \textit{HeteroFL} equal to 4-layer, 8-layer and 12-layer models in other baselines by reducing the hidden size from 768 to 456, 624 for weak and medium clients respectively.
For \textit{InclusiveFL}, we set the momentum factor $\beta=0.2$ for all GLUE benchmark and $\beta=0.5$ for medical NER datasets.

For GLUE dataset, we randomly sample a fraction of $0.02$ clients in each round  with $R=1000$.
For NER datasets, we conduct 5 local steps with batch size 64 in each communication round with $R=100$.
We set the same learning rate for each type of the model %
over all baselines.
We use FedAdam \cite{reddi2020adaptive} as the global aggregation protocol over homogeneous models for all baselines.

\begin{figure*}
     \centering
     \begin{subfigure}[b]{0.245\textwidth}
         \centering
         \includegraphics[trim=0 0 30 30,clip,width=\textwidth]{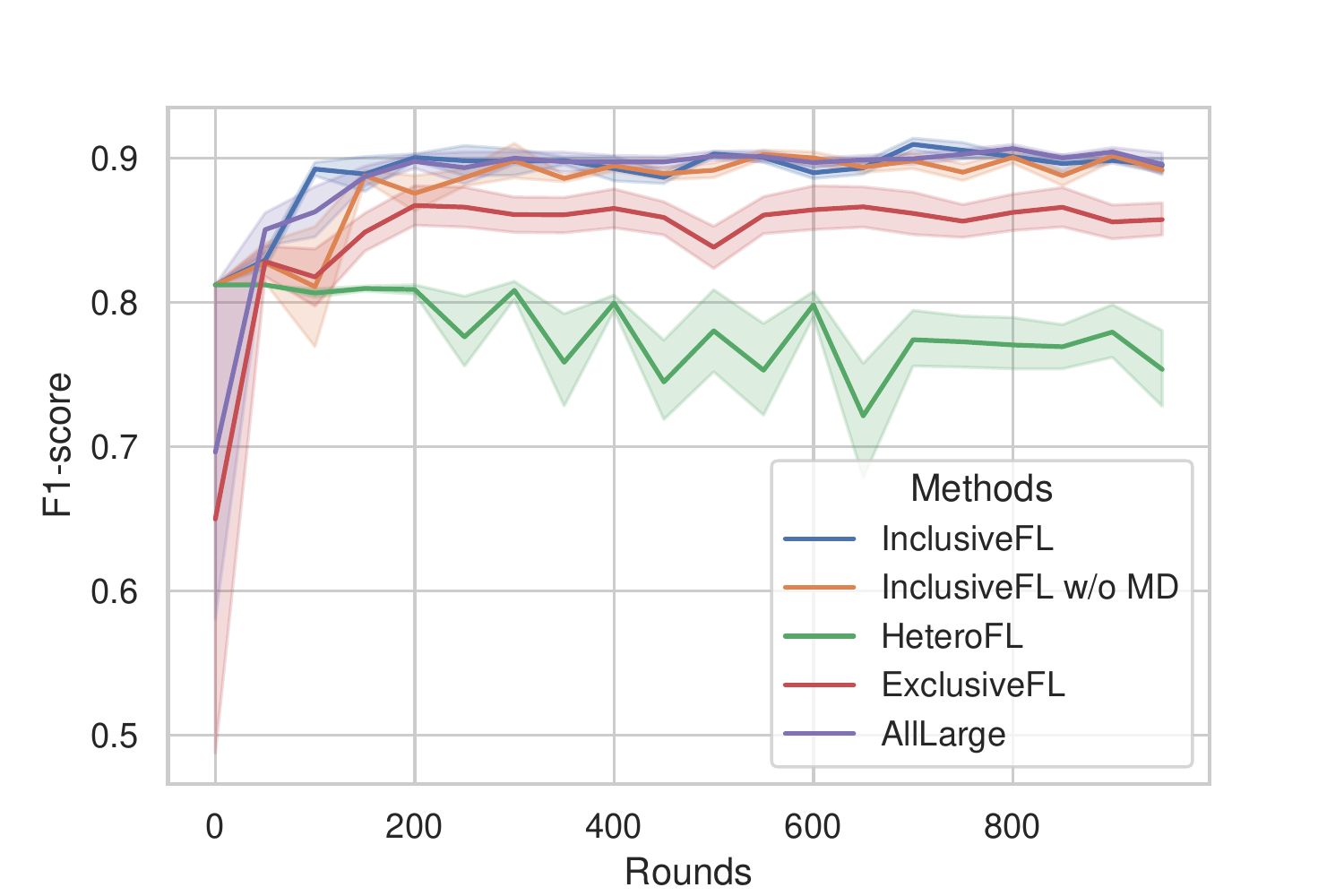}
         \caption{F1-score (MRPC).}
         \label{subfig-mrpc}
     \end{subfigure}
     \hfill
     \begin{subfigure}[b]{0.245\textwidth}
         \centering
        \includegraphics[trim=0 0 30 30,clip,width=\textwidth]{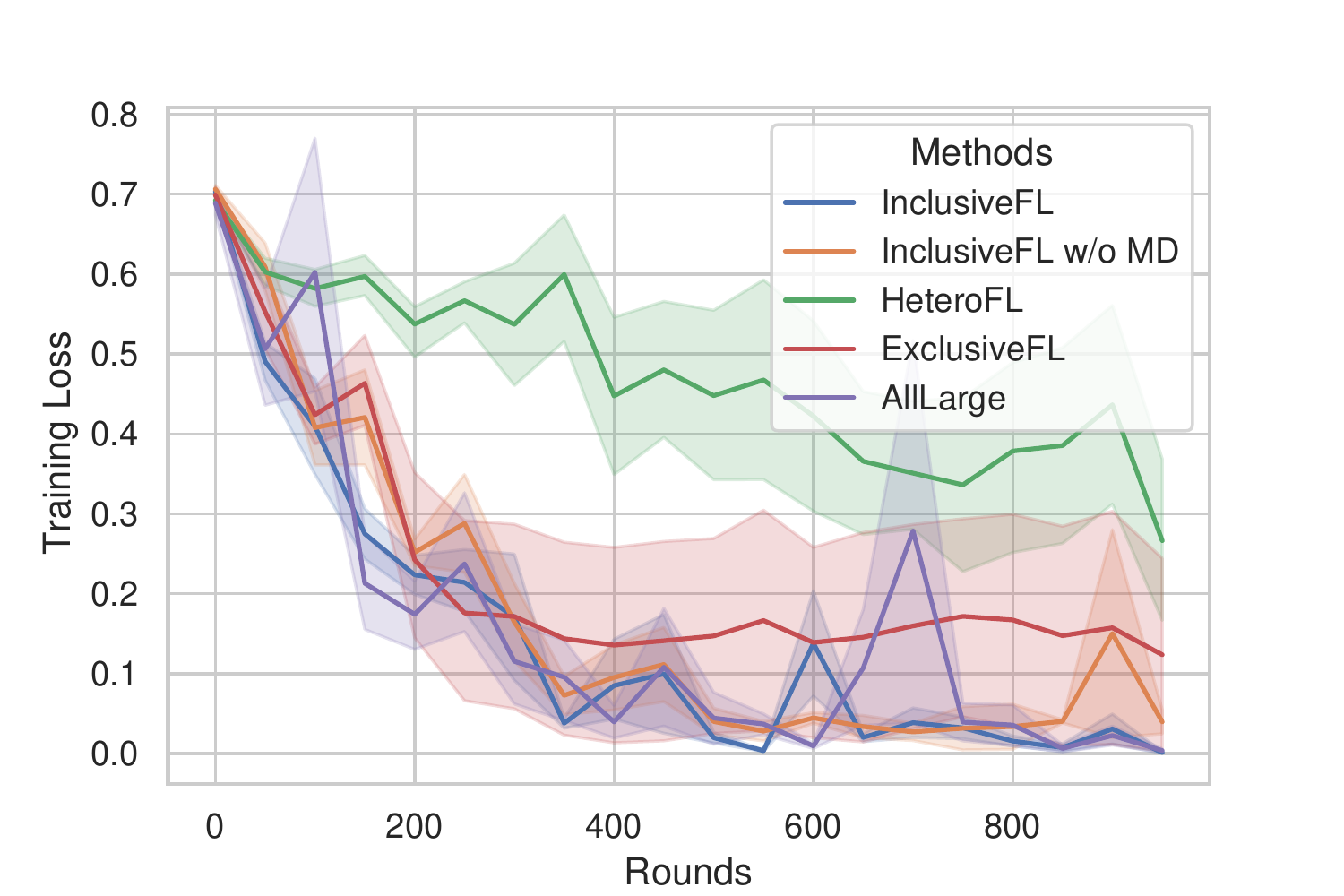}
         \caption{Training loss (MRPC).}
         \label{subfig-b}
     \end{subfigure}
     \hfill
     \begin{subfigure}[b]{0.245\textwidth}
         \centering
         \includegraphics[trim=0 0 30 30,clip,width=\textwidth]{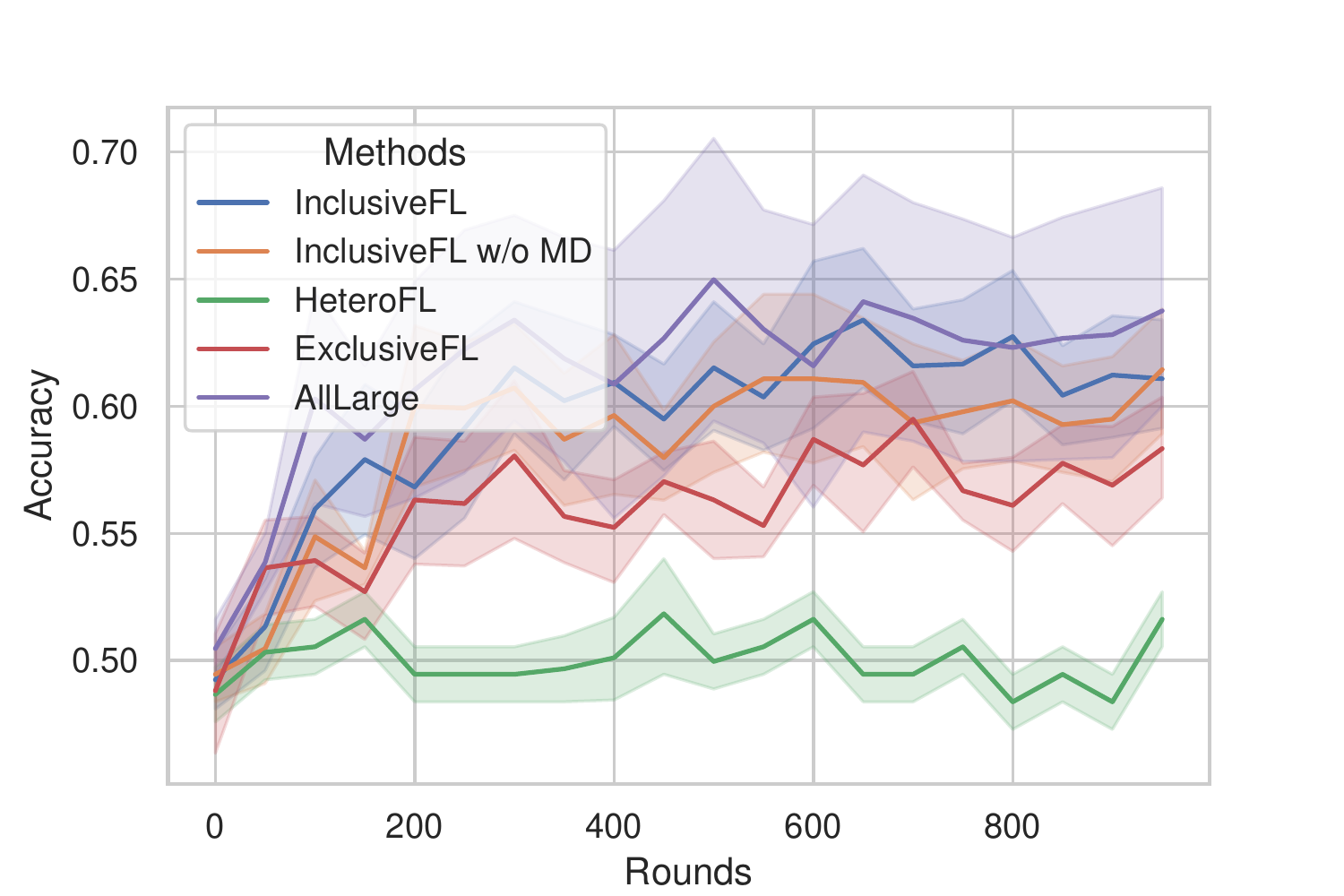}
         \caption{Accuracy (RTE).}
         \label{subfig-rte}
     \end{subfigure}
     \begin{subfigure}[b]{0.245\textwidth}
         \centering
         \includegraphics[trim=0 0 30 30,clip,width=\textwidth]{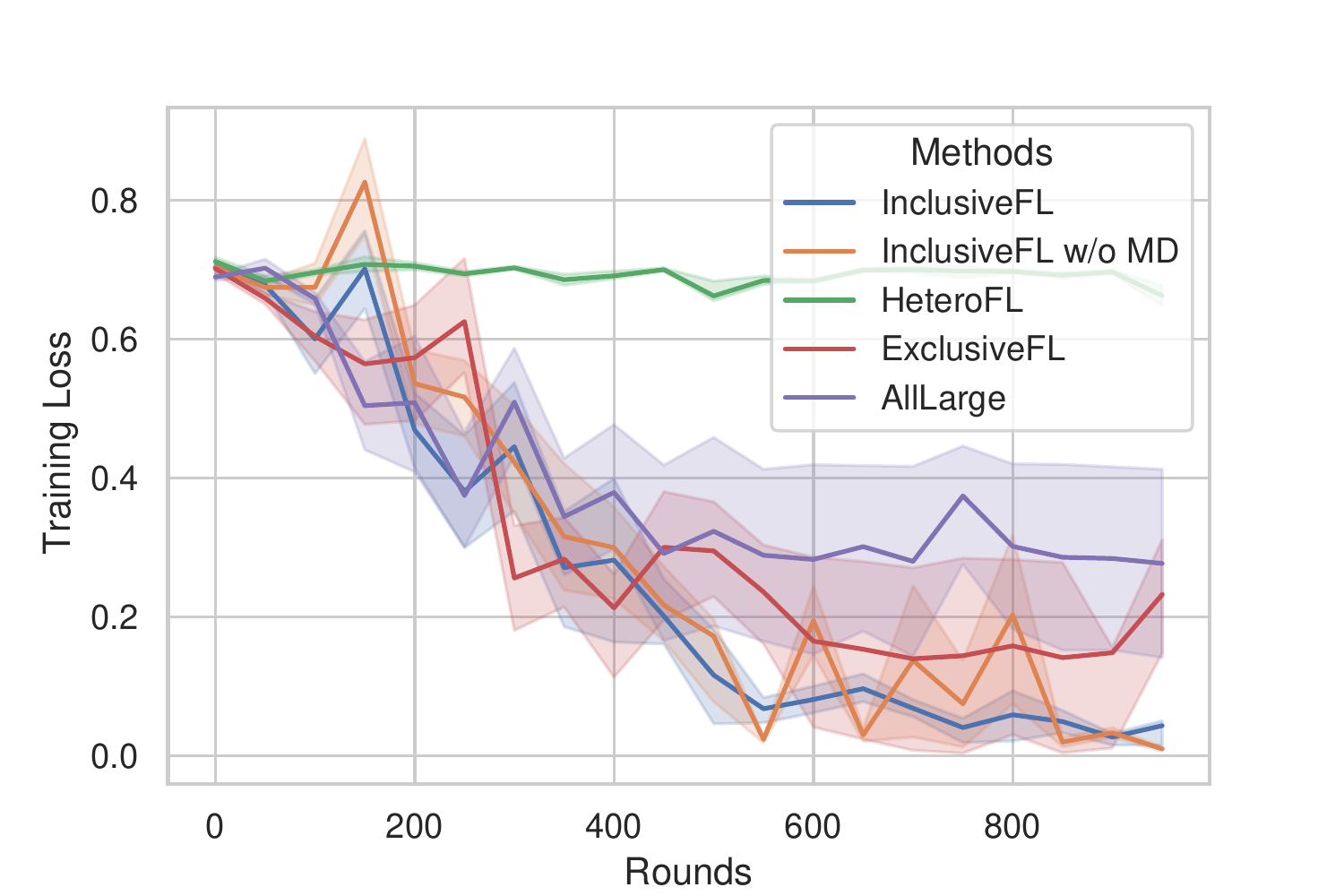}
         \caption{Training loss (RTE).}
         \label{subfig-d}
     \end{subfigure}
     
     \begin{subfigure}[b]{0.245\textwidth}
         \centering
         \includegraphics[trim=0 0 30 30,clip,width=\textwidth]{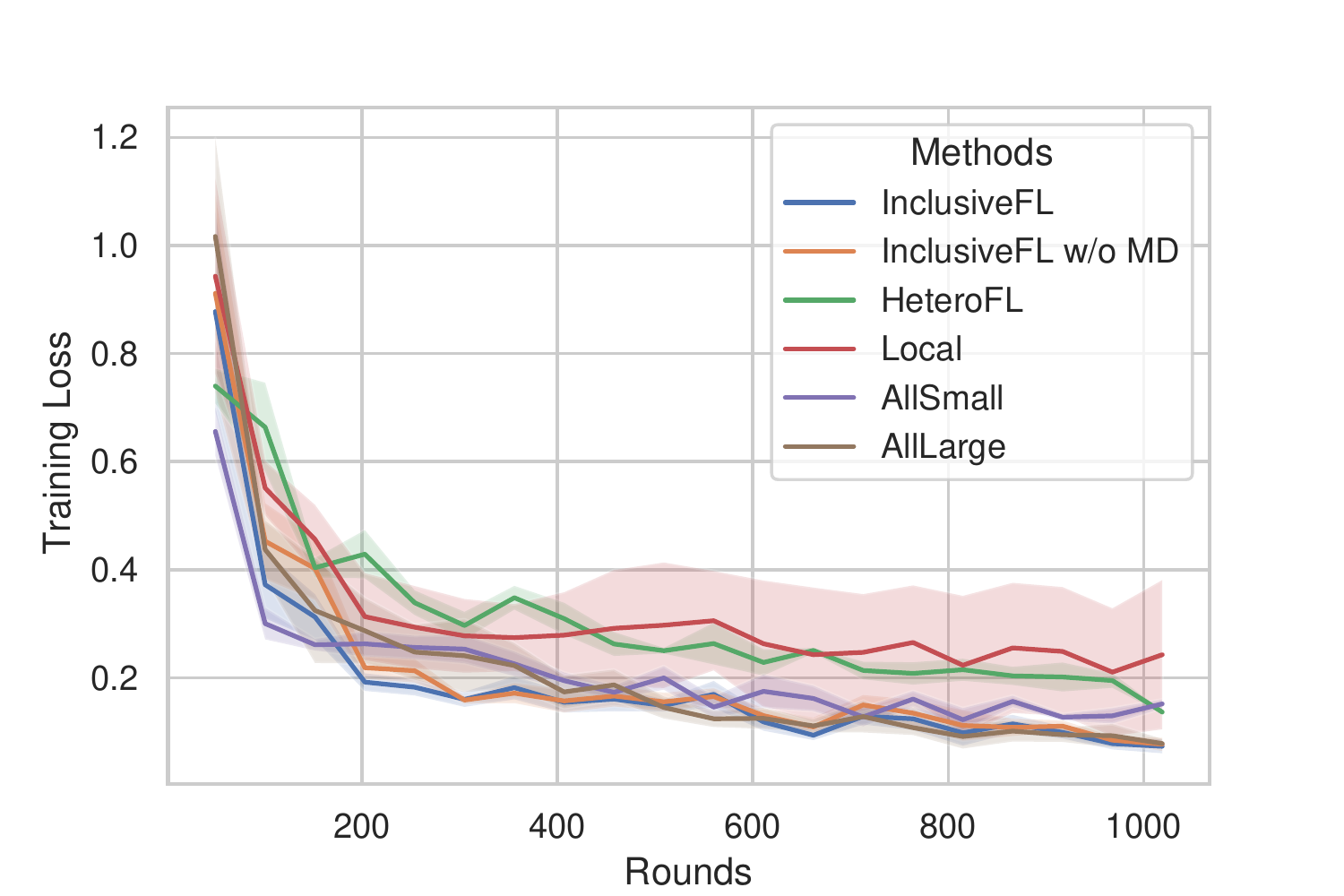}
         \caption{Training loss (CADEC).}
     \end{subfigure}
     \begin{subfigure}[b]{0.245\textwidth}
         \centering
         \includegraphics[trim=0 0 30 30,clip,width=\textwidth]{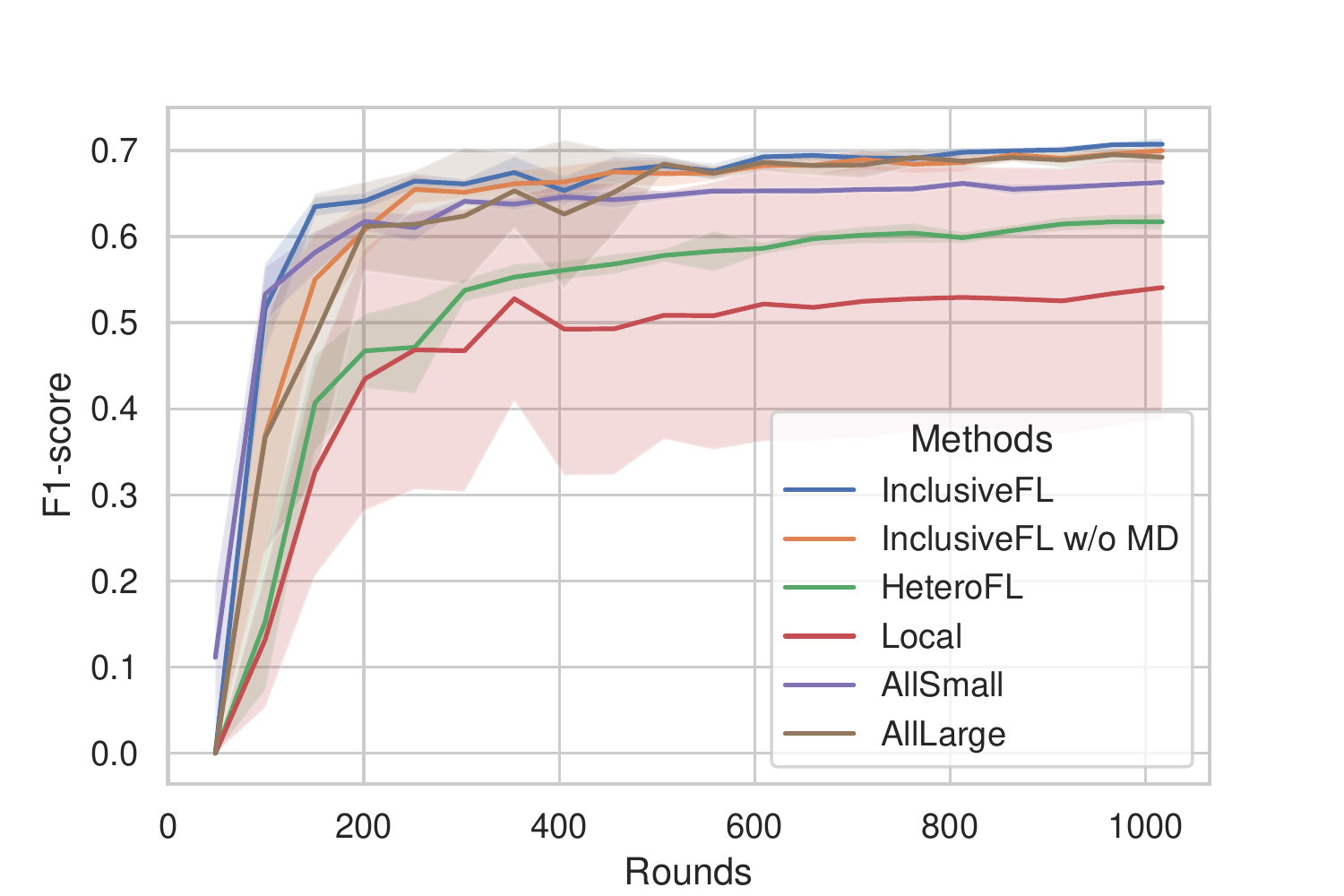}
         \caption{Overall F1-score (CADEC).}
         \label{subfig-cadec}
     \end{subfigure}
     \begin{subfigure}[b]{0.245\textwidth}
         \centering
         \includegraphics[trim=0 0 30 30,clip,width=\textwidth]{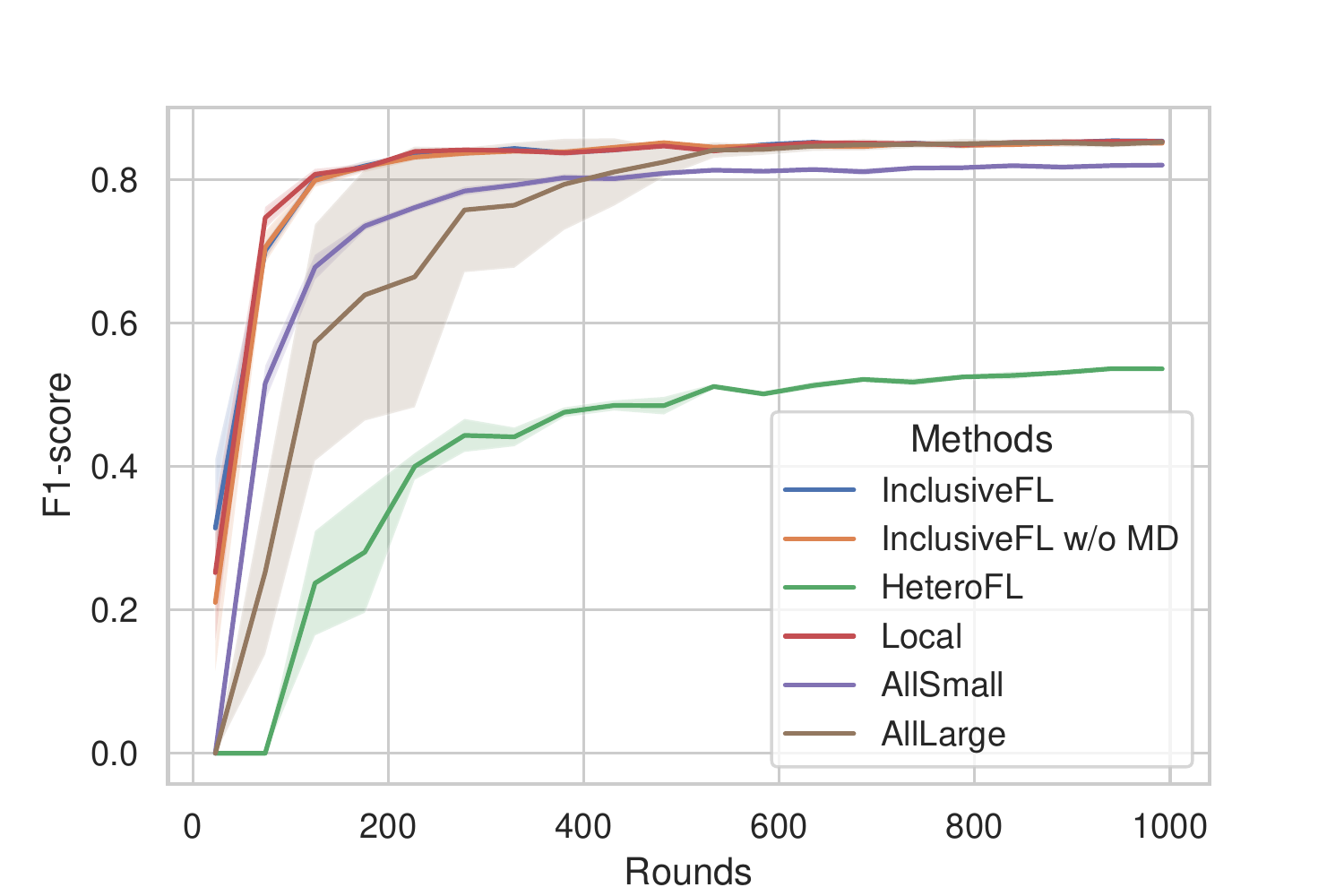}
         \caption{Overall F1-score (ADE).}
         \label{subfig-ade}
     \end{subfigure}
     \begin{subfigure}[b]{0.245\textwidth}
         \centering
         \includegraphics[trim=0 0 30 30,clip,width=\textwidth]{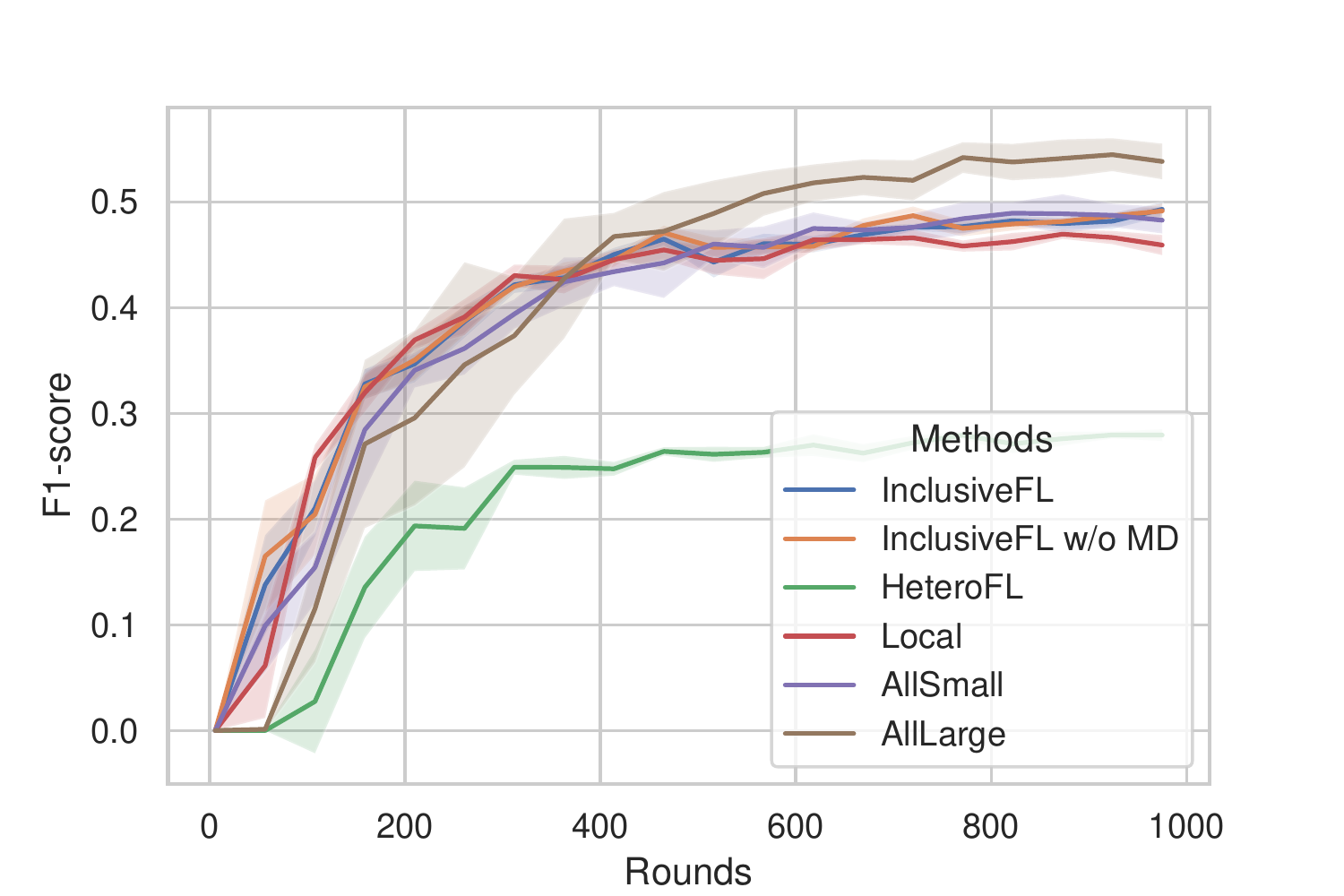}
         \caption{Overall F1-score (SMM4H).}
         \label{subfig-smm}
     \end{subfigure}
     \caption{Federated training convergence and best performance. Error band shows the standard error of 5 independent repeats.}
     \label{fig-conv}
\end{figure*}

\begin{table*}[h]
\centering
\caption{Averaged number of rounds for achieving the best performance in Table \ref{tab-glue}.}
\label{tab-rounds}
\begin{tabular}{c|ccccccccc}
\toprule[1.0pt]
Methods       & CoLA         & MNLI         & MRPC         & QNLI         & QQP          & RTE          & SST2         & STSB         & NER \\ \hline
InclusiveFL-w/o MD & 630          & 665          & 616          & 600          & 850          & 620          & 810          & 870          & 90  \\ %
InclusiveFL        & \textbf{580} & \textbf{570} & \textbf{515} & \textbf{585} & \textbf{740} & \textbf{510} & \textbf{520} & \textbf{805} & \textbf{75}  \\ 
\bottomrule[1.0pt]
\end{tabular}
\end{table*}

\begin{figure*}
    \centering
    \begin{subfigure}[b]{0.245\textwidth}
         \centering
        \includegraphics[trim=0 0 0 0,clip,width=\textwidth]{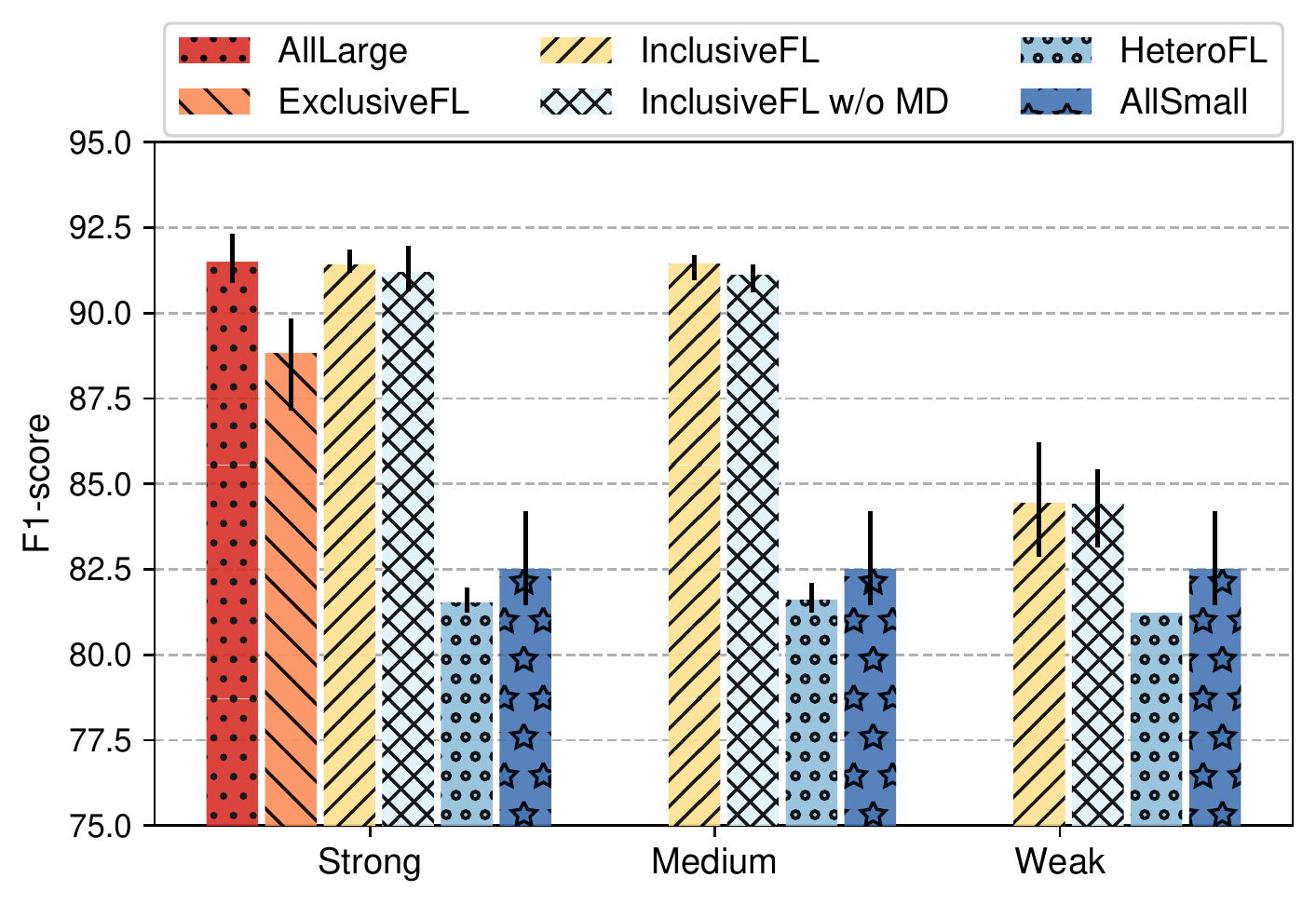}
         \caption{Averaged best F1-score (MRPC).}
     \end{subfigure}
    \begin{subfigure}[b]{0.245\textwidth}
         \centering
        \includegraphics[trim=0 0 0 0,clip,width=\textwidth]{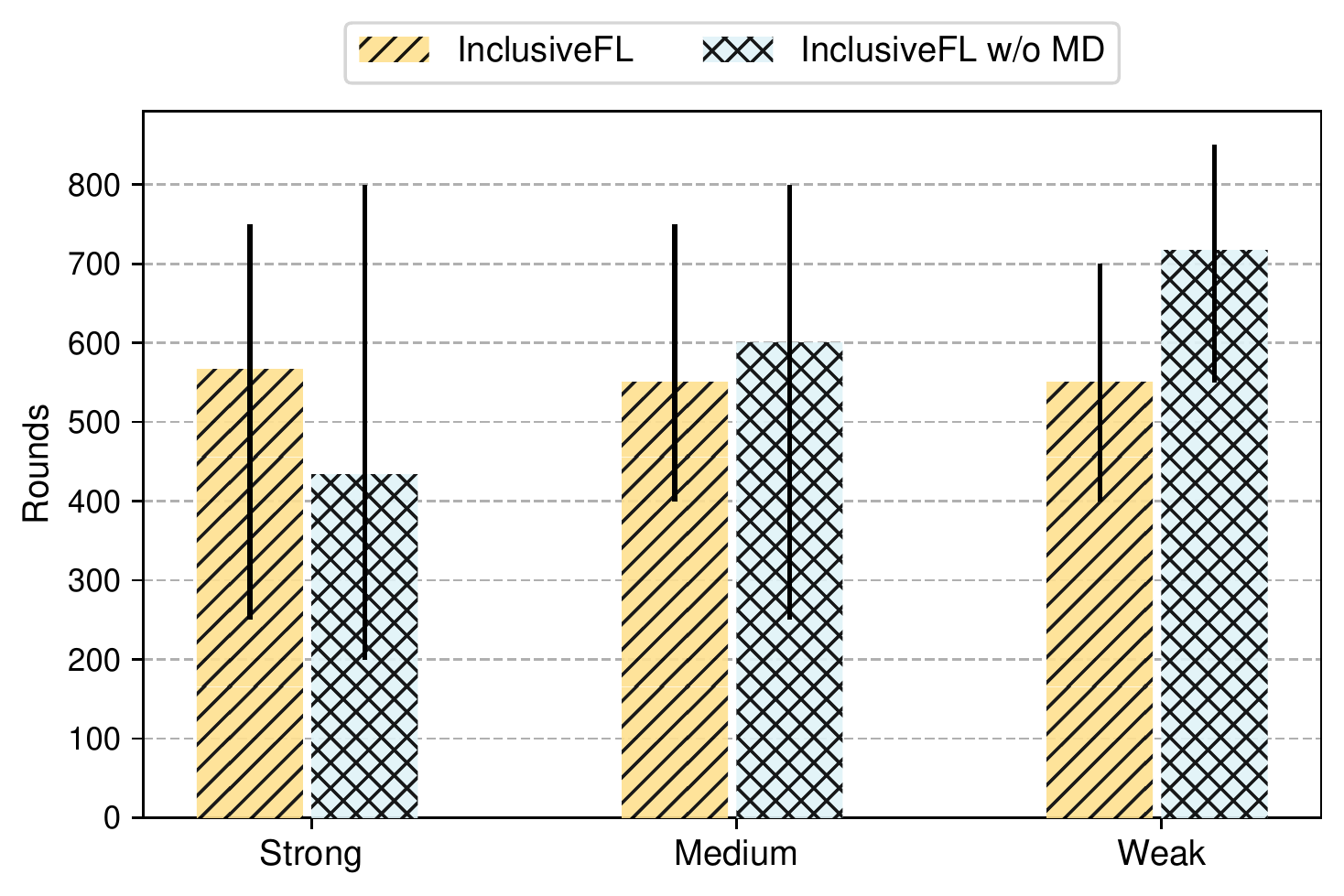}
         \caption{Averaged rounds (MRPC).}
     \end{subfigure}
    \begin{subfigure}[b]{0.245\textwidth}
         \centering
         \includegraphics[trim=0 0 0 0,clip,width=\textwidth]{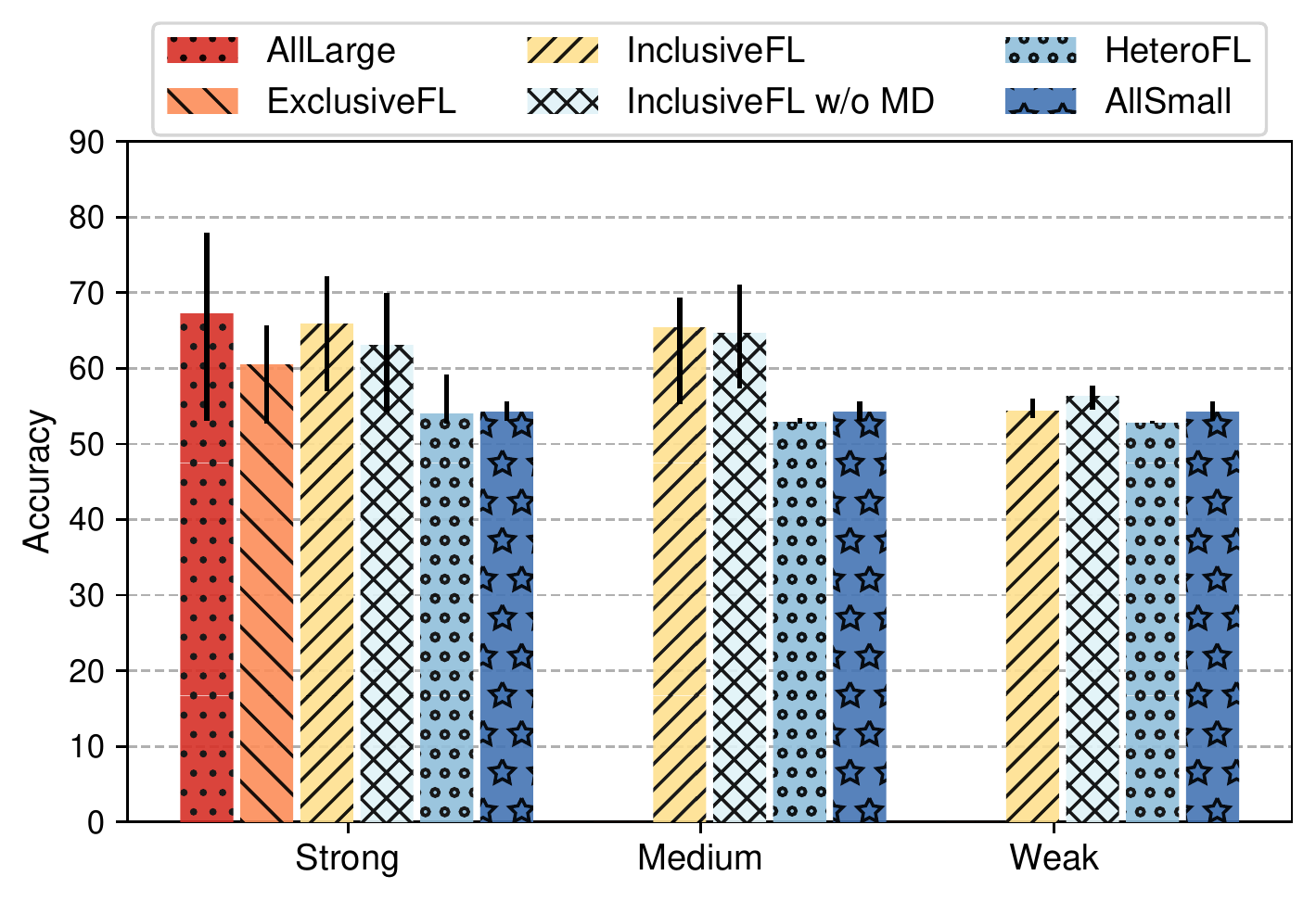}
         \caption{Averaged best accuracy (RTE).}
     \end{subfigure}
    \begin{subfigure}[b]{0.25\textwidth}
         \centering
        \includegraphics[trim=0 0 0 0,clip,width=\textwidth]{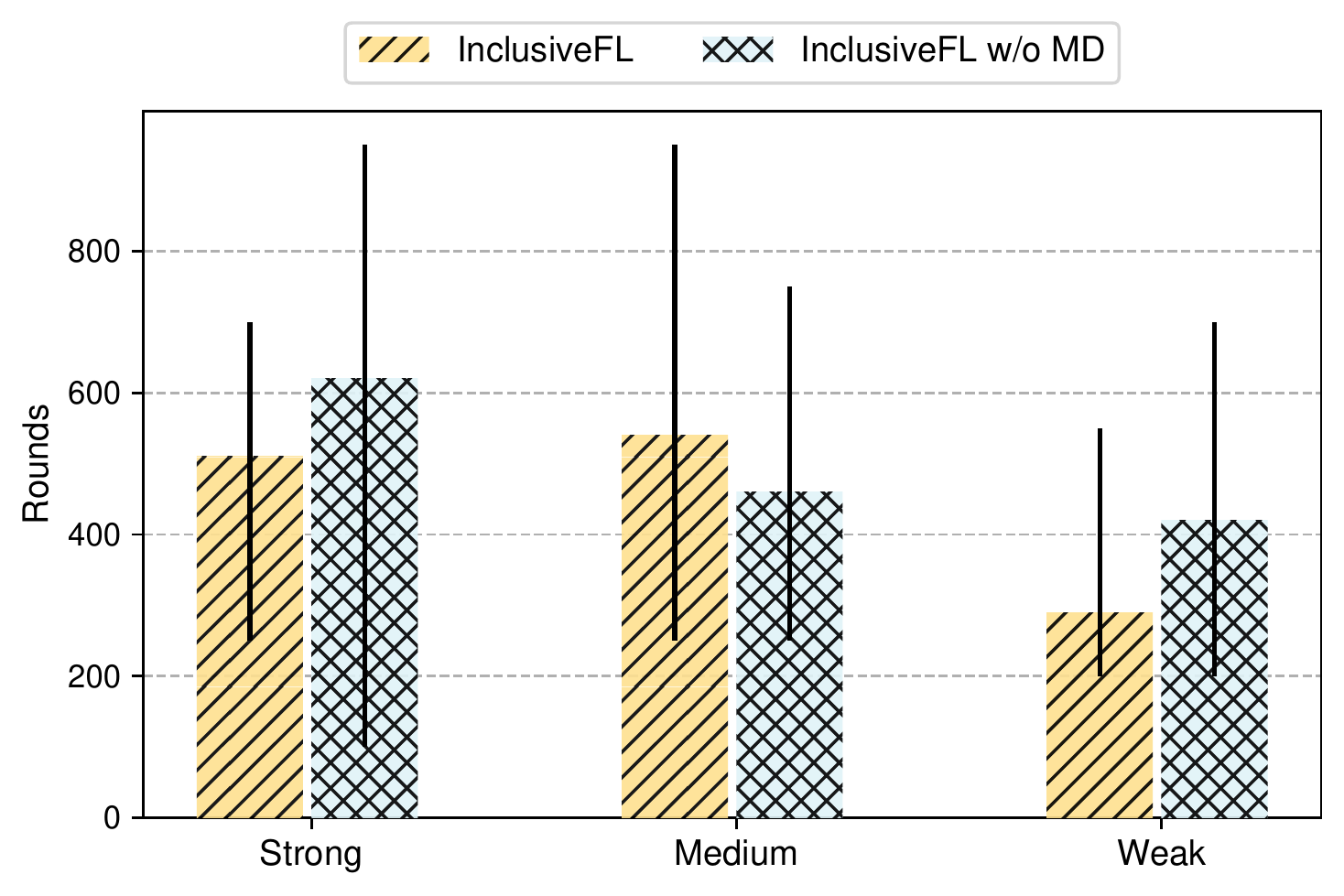}
         \caption{Averaged rounds (RTE).}
     \end{subfigure}
    \caption{Local inference performance in cross-device federated learning.}
    \label{fig-local-infer-glue}
\end{figure*}

\subsection{Performance Evaluation}
We evaluate the effectiveness of proposed methods by comparing the averaged best performance that each method can achieve after convergence within 1,000 training rounds.
Then we evaluate the efficiency and analyze the reason for improvement by showing the convergence curve and the average %
rounds %
required to achieve the best performance.

\textbf{Effectiveness.}
Results of IID setting on GLUE datasets and results of non-IID setting on medical NER datasets are shown in Table \ref{tab-glue} and \ref{tab-ner} respectively.
We observe that for all datasets, \textit{AllLarge} achieves the overall best performance, which indicates that training a large global model with all clients' data can achieve the optimal performance.
However, it is impractical for training participants with less resources to train large models locally.

For all other approaches that can be implemented across heterogeneous devices, our proposed approach \textit{InclusiveFL} achieves the best performance. 
It is reasonable because involving large models (\textit{InclusiveFL}) is better than only utilizing small models (\textit{AllSmall}) and training over all data (\textit{InclusiveFL}) is better than only training over partial data (\textit{ExclusiveFL} / \textit{Local}).
This observation also validates the effectiveness of our proposed momentum distillation.

We find that the most related work \textit{HeteroFL}~\cite{diao2020heterofl} performs worse than the naive baselines \textit{AllSmall} and \textit{ExclusiveFL} on all GLUE datasets.
And the averaged performance of \textit{HeteroFL} is also inferior to \textit{AllSmall} and \textit{ExclusiveFL} on medical datasets as shown in Table \ref{tab-ner}.
This  indicates that the method of sharing the top-left sub-matrix across heterogeneous devices in \textit{HeteroFL} may work for simple models with shallow layers (such as CNN in \cite{diao2020heterofl}) but not for more complex deep models (such as  12-layer RoBERTa).
Furthermore, we observe in Fig. \ref{fig-conv} that the starting performance of local models on weak and medium clients \textit{InclusiveFL} is better than \textit{HeteroFL}, which indicates that the small model in \textit{InclusiveFL} has a better parameter initialization than \textit{HeteroFL}.
Thus, one reason of the inferior utility is that cropping the network in \textit{HeteroFL} cannot fully utilized the pre-trained parameters while a layer-wise sharing in \textit{InclusiveFL} can better maintain the utility of a pre-trained model naturally.

In addition, we observe that \textit{ExclusiveFL} performs better than \textit{AllSmall} on GLUE benchmark and \textit{Local} performs worse than \textit{AllSmall}.
Thus, the performance gain from including more data and the gain from utilizing a large global model depends on the task and are hard to trade-off.
This observation enhances the contribution of \textit{InclusiveFL} for finding an effective way to aggregate over heterogeneous models and utilize performance gain from both inclusive data training and large models.

\textbf{Efficiency.}
By observing the convergence curve in Fig. \ref{fig-conv}, we find \textit{InclusiveFL} is approaching the performance of \textit{AllLarge} with a faster speed than \textit{InclusiveFL-w/o MD}, especially for the largest model with 12 layers.
For the IID setting as shown in Fig. \ref{subfig-mrpc}, \textit{InclusiveFL} achieves the F1-score of 89.43 within 400 rounds while \textit{InclusiveFL-w/o MD} requires 550 rounds.
For the non-IID setting in Fig. \ref{subfig-cadec}, \textit{InclusiveFL} achieves the accuracy of 60.5 within 650 while \textit{InclusiveFL-w/o MD} requires 950 rounds.
For all datasets, we list the averaged number of rounds %
required by \textit{InclusiveFL} with or without momentum distillation to achieve the best performance in Table \ref{tab-glue} and \ref{tab-ner}.
Thus, we can conclude that the proposed momentum distillation accelerates the performance of the largest global model by requiring less communication rounds.

\textbf{Discussion on the local model inference.}
To further investigate the reason for performance improvement, we observe the evaluation results on the local model for each group of clients.
For the cross-silo setting, local model inference is essential because each client has her own top layers of classifier for fitting local targets.
For the medium model in Fig. \ref{subfig-ade} and small model in \ref{subfig-smm}, we find the effect of boosting convergence is less obvious but still exists and leads to a higher performance in Table \ref{tab-ner} when the local model (8-layer for ADE and 4-layer for SMM4H) converges.

\begin{figure*}
    \centering
    \includegraphics[trim=0 0 0 0,clip,width=0.85\textwidth]{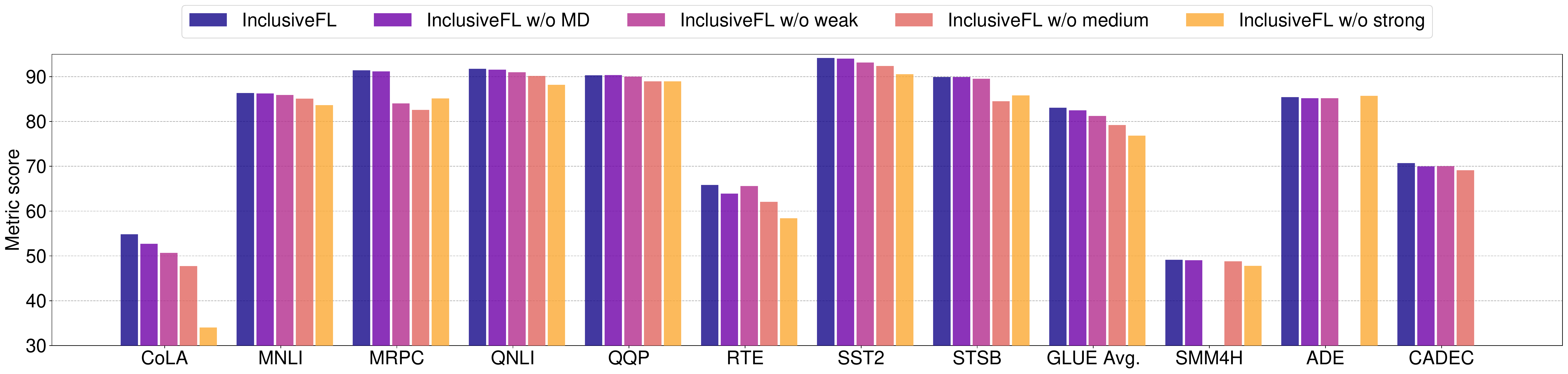}
    \caption{No one left behind: Ablation study of ignoring every single party for GLUE benchmark and medical NER datasets.}
    \label{fig-ab}
\end{figure*}

In the cross-device setting, the local model inference is also necessary when the inference is conducted on personal devices.
As shown in Fig. \ref{fig-local-infer-glue}, the final global model of \textit{ExclusiveFL} can only be deployed on devices of powerful clients while client-inclusive federated learning solutions (\textit{InclusiveFL}, \textit{HeteroFL}) and \textit{AllSmall} can be deployed on other clients with less storage and computation power.
We notice that \textit{InclusiveFL} performs best on each sub-global model, which validates the advantage of \textit{InclusiveFL} in local inference.
This also explains the performance advantage than another heterogeneous aggregation method \textit{HeteroFL} in a global inference scenario, because better sub-global models lay the foundation for a better large inference model.
In addition, the averaged rounds for achieving the best F1-score for MRPC and best accuracy for RTE is less for \textit{InclusiveFL} than \textit{InclusiveFL-w/o MD}, which shows the advantage of the proposed momentum distillation.

\begin{table}
\caption{Influence of different proportions of device types on the CoLA dataset.}
\label{tab-portion}
\begin{tabular}{c|ccc}
\toprule[1.0pt]
              & \textbf{1:1:1}   & \textbf{1:2:7}  & \textbf{7:2:1}  \\ \hline
AllLarge      & 63.03 & 63.03 & 63.03 \\
AllSmall      & 34.91 & 34.91 & 34.91 \\ \hline
ExclusiveFL      & 37.77   & 59.15  & \textbf{36.87}   \\ \hline
HeteroFL      & 8.15    & 9.60    & N/A      \\
InclusiveFL w/o MD & 52.69   & 60.35  & N/A  \\
InclusiveFL      & \textbf{54.85}   & \textbf{61.23}  & $34.96^*$  \\ 
\bottomrule[1.0pt]
\end{tabular}
\end{table}

\begin{figure}
    \centering
    \begin{subfigure}[b]{0.235\textwidth}
         \centering
         \includegraphics[trim=0 0 0 0,clip,width=\textwidth]{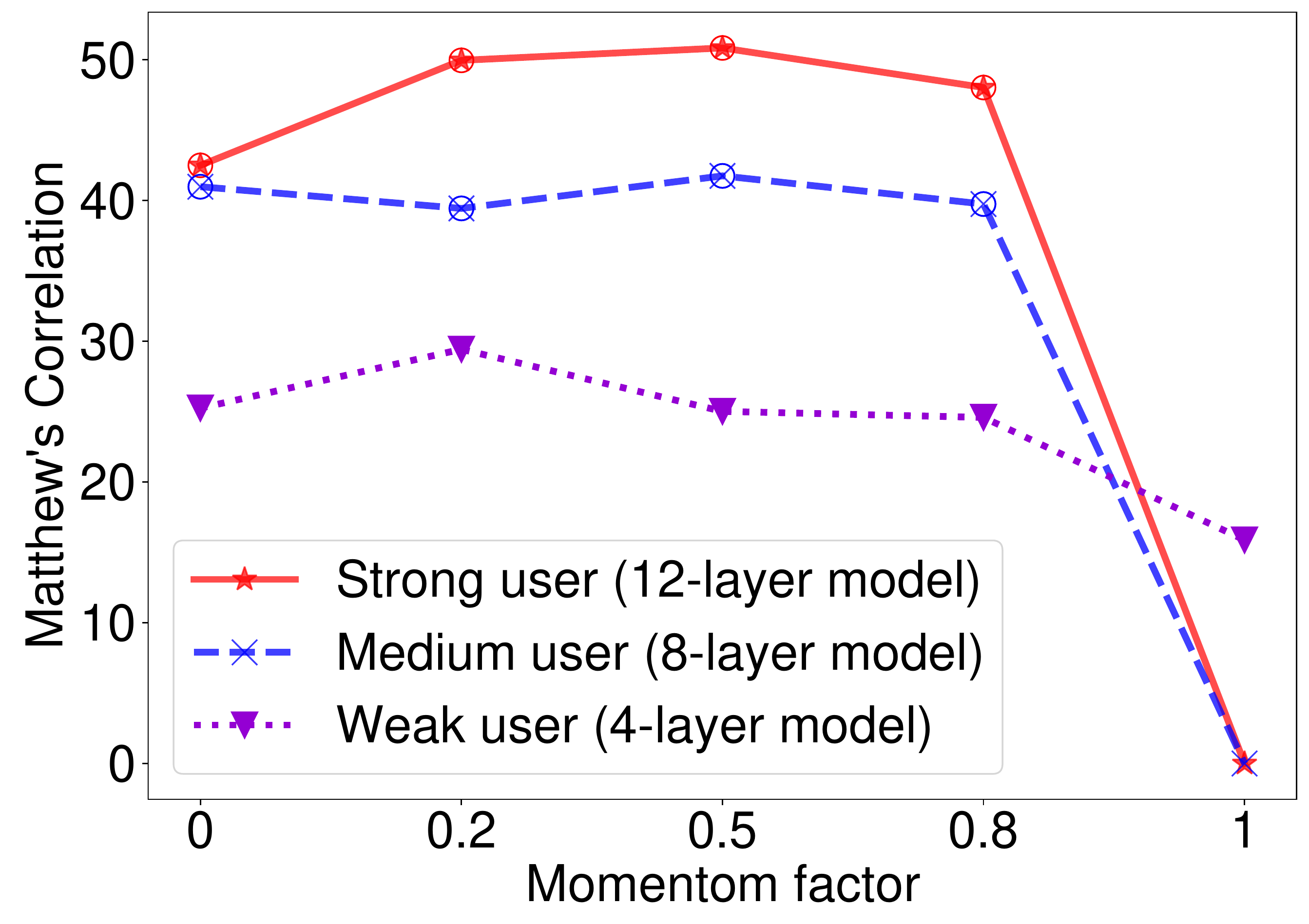}
         \caption{CoLA with \textit{InclusiveFL}$^*$.}
     \end{subfigure}
    \begin{subfigure}[b]{0.235\textwidth}
         \centering
         \includegraphics[trim=0 0 0 0,clip,width=\textwidth]{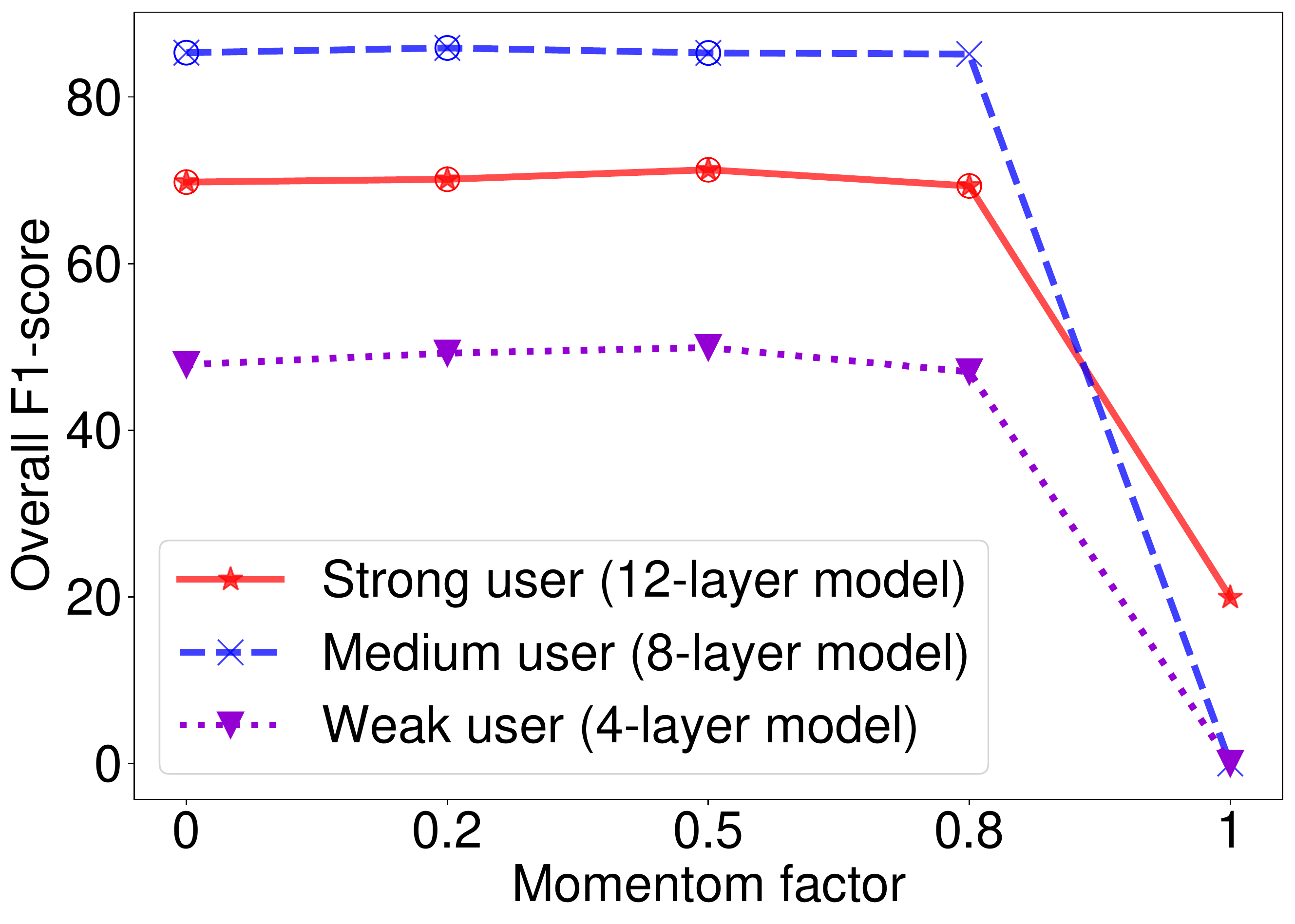}
         \caption{Medical NER with \textit{InclusiveFL}}
     \end{subfigure}
     \caption{Influence of momentum distillation $\beta$.}
     \label{fig-beta}
\end{figure}

\subsection{Ablation Study}
To validate the principle of ``No one left behind'', we conduct the ablation study in Fig. \ref{fig-ab} by ignoring each type of devices in heterogeneous federated learning on both GLUE benchmark and medical NER datasets.
For non-IID setting, \textit{InclusiveFL-w/o weak} for SMM4H, \textit{InclusiveFL-w/o medium} for ADE and \textit{InclusiveFL-w/o strong} for CADEC are not applicable because each client has different local token classification task and cannot be trained without the local dataset.
Among the three device types, we notice that \textit{InclusiveFL w/o strong} performs worse than \textit{InclusiveFL-w/o medium} and \textit{InclusiveFL-w/o weak}, which indicates that the contribution of powerful clients, medium clients and weak clients decrease progressively.
Powerful clients with more resources can learn knowledge with a larger model and contribute more.
We can observe that the client-inclusive federated learning (\textit{InclusiveFL} and \textit{InclusiveFL-w/o MD}) perform better than excluding each type of device, which validates our motivation that every single device should be included and contribute to federated training.

\subsection{Hyper-Parameter Analysis}
At last, we analyze two important hyper-parameters for \textit{InclusiveFL}.

\textbf{Device type proportion.}
To investigate the effectiveness of \textit{InclusiveFL} under different %
proportions of strong, medium, and weak devices, we present results with 1:2:7, 7:2:1 and a default setting of 1:1:1 on the CoLA dataset in Table \ref{tab-portion}.
In general, increasing the proportion of strong devices results in a better performance of \textit{ExclusiveFL} and \textit{HeteroFL}.
We can observe that the performance advantage of \textit{InclusiveFL} is consistent for two device proportions, compared with naive baselines and \textit{HeteroFL}.
But we notice that when the powerful client group is the minority (e.g., 7:2:1), the convergence of the largest model is slow and aggregating over heterogeneous devices is not stable.
Thus, we indicate one result in Table \ref{tab-portion} with a superscript of * when only adopt the momentum distillation without sharing layer-wise parameters across heterogeneous models.
And the performance is slightly worse than the best baseline \textit{ExclusiveFL} with a Matthew's correlation of 36.87.

\textbf{Momentum factor $\beta$.}
Then we tune the momentum factor $\beta$ of \textit{InclusiveFL} with various choices of $\{0, 0.2, 0.5, 0.8, 1\}$.
This parameter indicates the weight of gradient from larger model when applying the homophobic gradients to update each sub-global.
And $\beta=0$ equals the the version of \textit{InclusiveFL-w/o MD}.
We conduct evaluation on each type of sub-global model for the two cases: 
1) \textit{InclusiveFL}$^*$ that only applies the momentum distillation without heterogeneous parameter exchange on CoLA dataset in the IID cross-device setting.
2) \textit{InclusiveFL} on Medical NER dataset in non-IID cross-silo setting.
In Fig. \ref{fig-beta}, we indicate results that are superior than \textit{ExclusiveFL}, \textit{AllSmall} and \textit{HeteroFL} with the circle dot.
We notice that the optimal value of $\beta$ is around 0.5 for powerful clients under both settings.
The model on medium and weak devices also performs well when $\beta=0.2$ or $0.5$.
The performance of a $\beta>0.8$ may be ruined by injecting too much momentum from larger model to smaller model's gradients.
\section{Conclusions}
In this paper, we propose a client-inclusive framework \textit{InclusiveFL} for federated training over heterogeneous devices and remove the ideal assumption in conventional FL that all clients have sufficient device capability.
\textit{InclusiveFL} enables us to train a large global model with contributions from all clients by assigning models of different sizes to clients with different computing capabilities.
Thus, we avoid problems caused by ignoring weak clients or training with a small model.
Then, we propose an effective method to share the knowledge among multiple local models with different sizes by a layer-wise heterogeneous aggregation.
Besides, we propose a momentum knowledge distillation method for better transferring knowledge in big models of strong powerful clients to small models on weak clients.
Extensive experiments are conducted on benchmark data and real-world medical corpora with comparisons of both naive baselines and the existing state-of-the-art work under IID and Non-IID settings.
Results validate the accuracy improvement of \textit{InclusiveFL} on the large global model as well as smaller models when inference is conducted on local devices.

\begin{acks}
This work was supported by the National Natural Science Foundation of China under Grant numbers 62072460, 62076245, 62172424 and Beijing Natural Science Foundation under Grant number 4212022.
\end{acks}

\bibliographystyle{ACM-Reference-Format}
\bibliography{data/ref}

\end{document}